\documentclass[sigconf]{acmart}
\usepackage{caption}
\usepackage{subcaption}
\usepackage{algorithm}
\usepackage{algpseudocode}

\AtBeginDocument{%
  \providecommand\BibTeX{{%
    \normalfont B\kern-0.5em{\scshape i\kern-0.25em b}\kern-0.8em\TeX}}}

\begin{document}

%\title{Information theoretic point of view on hard crossover of neuroevolution}
%Suggestion from Marcus:
\title{Modularity based linkage model for neuroevolution}

\author{Yukai Qiao}
\authornote{Both authors contributed equally to this research.}

\orcid{0000-0002-9876-4353}
\affiliation{
    \institution{School of Information Technology and Electrical Engineering\\
      University of Queensland}
      \streetaddress{1 Th{\o}rv{\"a}ld Circle}
      \city{Brisbane}
      \country{Australia}
}
\email{kai.barnes@uqconnect.com}

\author{Marcus Gallagher}
\affiliation{
  \institution{School of Information Technology and Electrical Engineering\\
  University of Queensland}
  \streetaddress{1 Th{\o}rv{\"a}ld Circle}
  \city{Brisbane}
  \country{Australia}}
\email{marcusg@uq.edu.au}

\renewcommand{\shortauthors}{Yukai and Marcus}

\begin{abstract}
Crossover between neural networks is considered disruptive due to the strong functional dependency between connection weights. We propose a modularity-based linkage model at the weight level to preserve functionally dependent communities (building blocks) in neural networks during mixing. A proximity matrix is built by estimating the dependency between weights, then a community detection algorithm maximizing modularity is run on the graph described by such matrix. The resulting communities/groups of parameters are considered to be mutually independent and used as crossover masks in an optimal mixing EA. A variant is tested with an operator that neutralizes the permutation problem of neural networks to a degree. Experiments were performed on 8 and 10-bit parity problems as the intrinsic hierarchical nature of the dependencies in these problems are challenging to learn. The results show that our algorithm finds better, more functionally dependent linkage which leads to more successful crossover and better performance.
\end{abstract}

\begin{CCSXML}
<ccs2012>
   <concept>
       <concept_id>10003752.10003809.10003716.10011136.10011797.10011799</concept_id>
       <concept_desc>Theory of computation~Evolutionary algorithms</concept_desc>
       <concept_significance>500</concept_significance>
       </concept>
   <concept>
       <concept_id>10010147.10010257.10010293.10010294</concept_id>
       <concept_desc>Computing methodologies~Neural networks</concept_desc>
       <concept_significance>500</concept_significance>
       </concept>
 </ccs2012>
\end{CCSXML}

\ccsdesc[500]{Theory of computation~Evolutionary algorithms}
\ccsdesc[500]{Computing methodologies~Neural networks}

\keywords{neural networks, evolutionary algorithm, recombination, modularity, linkage model}

\maketitle

\section{Introduction}
The building block hypothesis states that Genetic Algorithms (GAs) work well by recombining small schemas of dependent genes into a better solution. The key to efficient recombination is the capability of preserving such schemas or so-called building blocks. Linkage Learning methods\cite{thierens2010linkage}\cite{thierens2011optimal}\cite{przewozniczek2020empirical} have shown success by estimating the dependencies between genes in order not to disrupt them during the recombination, thus improving the performance of the evolutionary process.

Previous research applying EAs on neural networks has focused on the so-called permutation problem \cite{hancock1992genetic}\cite{dragoni2014simba}\cite{stanley2002evolving}\cite{mahmood2007graph}\cite{thierens1993genetic}, which states that a neural network can have many configurations with the same functionality by a reordering of neurons (e.g. in a fully-connected layer). Recombination between two functionally similar neural networks thus does not always produce offspring that is functionally similar to its parents. A non-disruptive crossover can only then happen when the population is extremely converged in the parameter space which will hinder the search process. Solving the permutation problem would mean achieving permutation invariance from parameter space to functionality space and effectively reduce the search space in terms of the network architecture.

In addition to the permutation problem, in this paper we show that the disregard of inter-dependency between parameters is another primary source of disruption of recombination between neural networks. Neural networks typically implement a highly non-linear function, meaning that the optimization of their parameters can be a highly inseparable problem. The functionality of any weight influences, and is influenced by the perturbing of any other weight, as long as there is no disjoint part of the neural network. Performing recombination while preserving this interaction as much as possible while exchanging functional substructures is then a key factor in the success of the evolutionary process.

This issue has been considered in some previous work\cite{mouret2008mennag}\cite{radcliffe1993genetic} on indirect encoding neuroevolution. MENNAG\cite{mouret2008mennag} address this problem by encoding neural network as a generated graph of a Genetic Programming syntax tree, where connections between subgraphs represented by disjoint subtrees are naturally sparse, thus the overall generated neural network has a hierarchical modular structure. Swapping subtrees is then equivalent to swapping relatively independent substructures and such recombination was shown to be beneficial to performance by results in the paper and later empirical study\cite{qiao2020implementation}. To our knowledge, no efforts toward non-disruptive recombination have been made on fully connected neural networks with direct encoding. 

Linkage Learning attempts to model the dependency structure between the problem variables by observing the existing population. The best-known linkage model is the Linkage Tree (LT) model\cite{thierens2010linkage} where the mutual information between problem variables is estimated from the population and a tree-like linkage structure is learned by iteratively merging variables with the most mutual information. Linkage Learning does not require any domain knowledge of the problem, but this could lead to a problem (as pointed out in \cite{przewozniczek2020empirical}) in that it does not guarantee functional dependency of learned linkage as it simply captures correlated occurrences in the population. 

On the optimization of feedforward neural networks, we are not in a completely black-box situation, as we have a certain amount of domain knowledge of how parameters interact with each other. In this paper, we describe the Modularity-based Linkage model to exploit this fact. First, we build a proximity matrix where each element is the estimated first-order functional dependency between connection weights. A graph is then created by treating the proximity matrix as an adjacency matrix. Then, a community searching algorithm is applied to the graph to minimize the dependency between the searched communities where each community contains neural network weights that are deemed to be dependent within the community but independent to the rest of the neural network.

We carry out experiments on 8-bit parity and 10-bit parity problems since the non-separable nature of this problem enhances the inter-dependency of neural network parameter optimization, making it more challenging than it is. The results show that our algorithm finds better, more functionally dependent linkage which leads to more successful crossover and better performance compared with LT or when no crossover is applied. The results also indicate that solving either the permutation problem or the substructure disruption problem is not enough for efficient recombination, but both are required at the same time.

\section{Algorithm}
\subsection{Family Of Subsets}
Following the paradigm of linkage representation in \cite{thierens2011optimal}, a Family Of Subsets (FOS) $\mathcal F$ is a set of subsets of $\mathcal S =\{0,1,2,..,l-1\}$ which are the indices of problem parameters with a size of $l$. Each element $\mathcal F_{i}$ is assumed to be independent of the rest of the parameters $\mathcal S - \mathcal F_{i}$.

In a Marginal Product FOS, each element is mutually exclusive of the other; $\mathcal F_{i}\cap\mathcal F_{j}=\emptyset $ for any $\mathcal F_{i},\mathcal F_{j}\in\mathcal F$. In its simplest form, each subset consists of only one parameter which is the univariate FOS.
The Linkage Tree FOS describes a hierarchical dependency structure\cite{thierens2010linkage}. For any subset $\mathcal F_{i}$ with more than one element, there exist two subsets that are mutually exclusive, smaller in size and their union is exactly $\mathcal F_{i}$. The LT FOS can be seen as the result of a hierarchical clustering algorithm. At the start, we have the univariate FOS where each cluster contains one problem variable. New FOS elements are created by merging two existing FOS elements into a binary tree in a bottom-up fashion with each node being a subset of the FOS except the root. The whole Linkage Tree would contain $2l-1$ nodes which is also the size of the FOS. The size of the FOS can sometimes be a problem and since most of the bottom-leveled FOS elements do not describe valuable linkage information, a depth threshold is often applied while the linkage tree is being traversed.

\subsection{Optimal Mixing}
In Optimal Mixing (OM)\cite{thierens2011optimal} the FOS is traversed in a random order and an offspring is produced by two parents swapping genetic material using a different FOS element as a crossover mask. Specifically, the FOS element describes a subset of variables to be swapped into the first parent from the second parent. The offspring is only accepted if it outperforms the first parent and when it does it will replace the first parent. For a FOS with good quality, FOS elements are deemed to be independent to each other and OM would explore the outcome of all possible non-disruptive crossover learned by the linkage model.

For Recombinative Optimal Mixing, the same parents are used for each FOS element. For Gene-pool Optimal Mixing, the second parent is re-selected randomly each time a different FOS element is applied. These methods are summarized in Algorithm 1.
\begin{algorithm}
\caption{Optimal Mixing}\label{alg:om}
\begin{algorithmic}
\Function{ROM}{$p_{0},p_{1}, \mathcal F$}
\For{$\mathcal F_{i}$ in $\mathcal F$}
    \State $p_{0}^{new} \gets crossWithMask(p_{0},p_{1},\mathcal F_{i})$
    \State $evaluateFitness(p_{0}^{new})$
    \If{$p_{0}^{new}.fitness > p_{0}.fitness$}
        \State $p_{0} \gets p_{0}^{new}$
    \EndIf
\EndFor
\State Return $D$
\EndFunction
\\
\Function{GOM}{$p_{0},P, \mathcal F$}
\For{$\mathcal F_{i}$ in $\mathcal F$}
    \State $p_{1} \gets randomSelection(P-p_{0})$
    \State $p_{0}^{new} \gets crossWithMask(p_{0},p_{1},\mathcal F_{i})$
    \State $evaluateFitness(p_{0}^{new})$
    \If{$p_{0}^{new}.fitness > p_{0}.fitness$}
        \State $p_{0} \gets p_{0}^{new}$
    \EndIf
\EndFor
\State Return $D$
\EndFunction
\end{algorithmic}
\end{algorithm}

\subsection{Neuron Similarity}
A big issue with recombination in neuroevolution is the permutation problem. A neural network can have many juxtaposes with the same functionality but different orderings of the neurons in the same layers. This means mixing between functionally close neural networks doesn't always result in functionally close offspring. To attempt to neutralize this to some degree, we implemented the Neuron Similarity operator inspired by \cite{dragoni2014simba}. 

In this operator, the activation (output) values of the neurons of each individual are recorded for all inputs. During recombination, the neurons of the second parent are rearranged temporarily by comparing their activation values to the first parent. In each layer, each neuron is matched to a neuron of the first parent where the sum of the differences of the activation values over a batch of input is minimal, in a greedy manner. This operator is summarized in Algorithm 2.

\begin{algorithm}
\caption{Neuron Similarity}\label{alg:ns}
\begin{algorithmic}
\Function{neuronSimilarityRearrange}{$p_{0},p_{1}$}
    \State $a^{0}\gets p_{0}.activationValues$
    \State $a^{1}\gets p_{1}.activationValues$
    \For{$l=1$ to $|p_{0}.hiddenLayers|$}
        \State $R\gets$ $\{1,2,...|p_{0}.hiddenLayers.neurons|\}$
        \For{$j=1$ to $|p_{1}.hiddenLayers.neurons|$}
            \For{$i$ in $R$}
                \State $D_{i}\gets L1Distance(a^{0}_{li},a^{1}_{li})$
            \EndFor
            \State $perm_{j} \gets argMin(D)$
            \State $R\gets R-argMin(D)$
        \EndFor
        \State $applyPermutation(p_{1}.hiddenLayers[l],perm)$
        
    \EndFor
\State Return $p_{1}$
\EndFunction
\end{algorithmic}
\end{algorithm}

\subsection{Modularity-based linkage model}

Our proposed modularity-based linkage model attempts to preserve interaction between weights by finding a modular decomposition of a graph, where each \textit{vertex} is a weight in the neural network and each edge represents the estimated strength of the interaction between the two vertices connected to the edge. 

Although a fully connected neural network is non-separable in terms of dependency between the parameters, we only consider two weights to be 'connected' when one weight connects to a neuron and the other weight connects from the same neuron. The strength of interaction of the said pair of weights is estimated to be the absolute value of the product of the weights. This provides a crude approximation as it ignores the non-linearity of the activation function and the effect of other inputs of the neuron. We define a weights proximity matrix as a matrix of interaction values between all pairs of weights in each layer of a network. Algorithm 3 summarizes the calculation of this matrix.

\begin{algorithm}
\caption{Weights Proximity Matrix}\label{alg:matrix}
\begin{algorithmic}
\Function{weightProximity}{$L,N,E,w$}
\State $L \gets$ layer number of the neural network
\State $N_{l} \gets$ neuron number of the $l^{th}$ layer 
\State $E \gets$ all weights of the neural network
\State $w_{l,i,j} \gets$ weight connecting from the $i^{th}$ neuron of layer $l$ to the $j^{th}$ neuron of layer $l+1$
\State $D \gets E\times E \rightarrow 0 \gets$ weight dependency matrix initialized with value 0 
\For{$l =0 $ to $L - 2$}
    \For{$i = 0$ to $N_{l}$}
        \For{$j = 0$ to $N_{l+1}$}
            \For{$k = 0$ to $N_{l+2}$}
                \State $D(w_{l,i,j},w_{l+1,j,k}) \gets abs(w_{l,i,j}w_{l+1,j,k})$
            \EndFor
        \EndFor
    \EndFor
\EndFor
\State Return $D$
\EndFunction
\end{algorithmic}
\end{algorithm}

\begin{figure}[h]
  \centering
  \includegraphics[width=\linewidth]{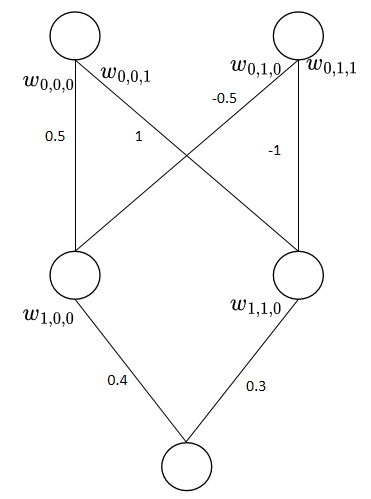}
  \caption{An example neural network}
\end{figure}

\begin{table}[h!]
\centering
\begin{tabular}{ |c|c|c|c|c|c|c| } 

 \hline
       & $w_{0,0,0}$ & $w_{0,0,1}$ & $w_{0,1,0}$ & $w_{0,1,1}$ & $w_{1,0,0}$ & $w_{1,1,0}$ \\ 
 \hline
 $w_{0,0,0}$ & 0 & 0 & 0 & 0 & 0.2 & 0 \\ 
 \hline
 $w_{0,0,1}$ & 0 & 0 & 0 & 0 & 0 & 0.3  \\
 \hline
 $w_{0,1,0}$ & 0 & 0 & 0 & 0 & 0.2 & 0  \\
 \hline
 $w_{0,1,1}$ & 0 & 0 & 0 & 0 & 0 & 0.3  \\
 \hline
 $w_{1,0,0}$ & 0.2 & 0 & 0.2 & 0 & 0 & 0  \\
 \hline
 $w_{1,1,0}$ & 0 & 0.3 & 0 & 0.3 & 0 & 0  \\
 \hline
\end{tabular}
\caption{Corresponding proximity matrix of Figure.1}
\end{table}

The recalculation of the proximity matrix after merging groups can be done easily with mutual information-based metrics due to the grouping property\cite{thierens2010linkage}, making hierarchical clustering a suitable method of estimating the dependency structure. In order to better estimate the dependencies within a group, we apply the concept of modularity: for a given graph and the grouping of its vertices, modularity describes how sparse the edges are between groups than within groups. By treating the proximity matrix as an adjacency matrix of a graph, finding a grouping that maximizes modularity would be close to finding subsets of highly dependent variables while keeping them independent between different subsets, which fits the purpose of FOS.

For two groups,modularity $Q$ is defined as\cite{newman2006modularity}:
\begin{equation}
Q=\frac{1}{4m}\sum_{ij}{(A_{ij}-\frac{k_{i}k_{j}}{2m})s_{i}s_{j}}
\end{equation}
where $A_{ij}$ is the number of edges between vertices $i$ and $j$, $k_{i}$ is the degree of vertex $i$, $m$ is the total number of edges, $s_{i}$ indicates the group $i$ belongs to, where a value of 1 indicates group 1 and a value of -1 indicates group 2. This formula can be generalized to positive weighted graphs, where the number of edges would represent weights  and the total number of edges would represent the sum of weights instead.

The recombination operator requires two individuals, however the aforementioned metric only describes the proximity structure of a single entity. This is the main reason a between-group average hierarchical clustering method like UPGMA as implemented in GOMEA\cite{bosman2012linkage}, cannot be used directly since we cannot apply this dependency measurement to a population. 

The definition of modularity can be shown to be linear with graphs having the same number of vertices and the same sum of edges. Let adjacency matrix $A_{n}$ describe a graph calculated from individual $n$. Assuming our goal is to find a grouping maximizing modularity for two individuals simultaneously $Q_{A_{1}}+Q_{A_{2}}$, assuming we have a fixed architecture, the sizes of the networks are the same, we have:
\begin{equation}
\begin{split}
    Q_{A_{1}} + Q_{A_{2}} & =\frac{1}{4m_{1}}\sum_{ij}{(A_{1ij}-\frac{k_{1i}k_{1j}}{2m_{1}})s_{1i}s_{1j}}\\
                          & + \frac{1}{4m_{2}}\sum_{ij}{(A_{2ij}-\frac{k_{2i}k_{2j}}{2m_{2}})s_{2i}s_{2j}}\\
\end{split}
\end{equation}
Since we are finding a single grouping for both graphs simultaneously, $s_{1}$ will be equal to $s_{2}$. By normalizing $A_{2}$ with a factor of $\frac{m_{1}}{m_{2}}$ we can get:
\begin{equation}
\begin{split}
    Q_{A_{1}} + Q_{\frac{m_{1}}{m_{2}}A_{2}} & =\frac{1}{4m_{1}}\sum_{ij}{(A_{1ij}-\frac{k_{1i}k_{1j}}{2m_{1}})s_{1i}s_{1j}}\\
    & + \frac{1}{4m_{1}}\sum_{ij}{(\frac{m_{1}}{m_{2}}A_{2ij}-\frac{\frac{m_{1}}{m_{2}}k_{2i}\frac{m_{1}}{m_{2}}k_{2j}}{2m_{1}})s_{1i}s_{1j}}\\
    & =2Q_{A_{1} + \frac{m_{1}}{m_{2}}A_{2}}\\
\end{split}
\end{equation}
That is, to maximize the modularity of a graph described by adjacency matrix $(A_{1}+\frac{m_{1}}{m_{2}}A_{2})$ is to find a single grouping maximizing a linear combination of the modularity of graph 1 and graph 2. With a common assumption that every individual of the population is from the same distribution, the value of $\frac{m_{1}}{m_{2}}$ should be close to 1. This means the modular decomposition should not lean too much to one side where only one individual's substructures are being preserved. The linear nature of this approximation also indicates that it is not suitable to be generalized to multiple individuals as the substructures of some individuals are deemed to be ignored. This leads to the fact that a population-wise linkage structure cannot be learned effectively and thus a pair-wise linkage structure must be learned before crossover. Another consequence is that Gene-Pool optimal mixing will not be compatible with this linkage model anymore since changing the second donor would mean recalculating the linkage and the computational overhead would be unacceptable.

We apply the Leiden algorithm to optimize the community decomposition of the graph described by the proximity matrix as it arguably the state of the art algorithm for this task\cite{traag2019louvain}. Empirical results\cite{traag2019louvain} showed that it runs in near linear time in the number of edges, this is exceptionally important as our proximity matrix is very sparse and the graph can be represented as an adjacency list instead. A downside, however, is that the algorithm does not provide a hierarchical structure of the communities, unlike Newman's algorithm\cite{newman2006modularity}. This means we cannot build a linkage tree but have to use the marginal product model. Newman's algorithm, however, requires eigen-decomposition of the dependency matrix and it would simply not scale when there are more than a few hundred of weights with a time complexity of $O(n^{3})$. Time complexity is also a reason for not using UPGMA as it runs in $O(n^{2})$\cite{gronau2007optimal}.

Our modularity-based linkage EA is summarized in Algorithm 4.

\begin{algorithm}
\caption{Modularity-based Linkage model EA}\label{alg:mod}
\begin{algorithmic}
\State $P \gets$ Initialize population and evaluated

\While{not terminationConditionMet}
    \For {$i = 1$ to $\lceil elitismRate \times |P|\rceil$}
        \State $P_{i}^{next} \gets P_{i}$
    \EndFor
    \While{$|P^{next}|<|P|$}
        \State $p_{0} \gets rankProportionalSelection(P)$
        \State $p_{1} \gets copy(randomSelection(P))$
        \If{$neuronSimilarity$}
            \State $p_{1} \gets neuronSimilarityRearrange(p_{0},p_{1})$
        \EndIf
        \State $w_{0} \gets p_{0}.weights$
        \State $w_{1} \gets p_{1}.weights$
        \State $w_{1} \gets \frac{sum(w_{0})}{sum(w_{1})}w_{1}$
        \State $G \gets adjacencyList(weightProximity(\frac{w_{0}+w_{1}}{2}))$
        \State $\mathcal F \gets LeidenCommunityDetection(G)$
        \State $P^{next} \gets P^{next} + ROM(p_{0}, p_{1}, \mathcal F)$
    \EndWhile
    \State $P \gets P^{next}$
\EndWhile
\State Return $D$
\end{algorithmic}
\end{algorithm}

\section{Experiments}

\subsection{Parity Problem}
In nature, fully connected feed-forward neural networks have overlapping building blocks: it is hard to find a sub-network that does not interact with any other sub-network. To further enhance this property, we choose the parity problem to test the capability of finding good linkages with our algorithm. 

The input of the parity problem is an $n$-bit bitstring, and the target output is the parity of the sum of 1s in the bitstring. This problem cannot be decomposed into a linear combination of sub-problems meaning that any building block found will be overlapping with at least one other. Even if the problem is solved optimally by building a hierarchical XOR function the solution will consist of building blocks with high inter-dependency.

%20 experiments are run on both 8-bit parity and 10-bit parity problem.
\subsection{Setup}
On each problem, each experiment is run for 20 repeated trials.
For the 8-bit parity problem, the size of the layers of neural networks including input and output layers is:[8,8,8,8,1] with three hidden layers. For the 10-bit parity problem it is [10,10,10,10,1]. All neurons have $tanh()$ as the activation function. Weight initialization is Gaussian with a mean of 0 and a standard deviation of 3. The mutation is performed by adding a Gaussian perturbation with a standard deviation of 0.2. The mutation rate is 0.3. The experiments stop after a maximum of $10^6$ evaluations. The first parent is selected by rank-proportional selection. The population size is 100 with an elitism rate of 0.01.

To make a fair comparison to recombination without linkage and mutation-only reproduction, we apply the idea of only accepting successful offspring to setups that do not require FOS in order to give them a similar amount of pressure in the reproduction phase. As a matter of fact, such an operator improves the performance significantly, making our results more convincing. The different algorithmic setups tested are as follows:
\begin{description}
   
   \item[MOD] Recombinative optimal mixing, Modularity-based linkage model. 
   \item[MOD\_NS] Same as MOD but the neuron similarity operator is applied to the second parent to align its neurons to the first parent.
   \item[Uniform] Only accept successful reproduction, Uniform crossover. In Uniform crossover the offspring inherits each parameter from both parents with equal probability.
   \item[Uniform\_NS] Same as Uniform but the neuron similarity operator is applied to the second parent to align its neurons to the first parent.
   \item[No] Only accept successful reproduction, no crossover. This serves as a baseline since if recombination is beneficial, the performance should be at least as good as no crossover at all.
   \item[LT] Gene-pool optimal mixing, Mutual information-based Linkage Tree model. The focus of this work is to compare the quality of linkage found in neural networks, thus we only implemented a minimal version of GOMEA adapted to real value with no covariance adaptation mechanism\cite{bouter2017exploiting}. There is no neuron similarity variant since the linkage tree must be built on mutual information calculated from the whole population. Rearranging neurons before crossover would invalidate the linkages learned beforehand.
    
\end{description}

\subsection{Metrics}
To evaluate the results of our experiments, the following metrics are used (averages reported over the 20 runs):
%All of the metrics are averaged over 20 experiments.
\begin{description}
   
   \item[Fitness] The fitness is the accuracy of the prediction of parity given the bit-string. A negative output of the neural network is treated as 0 and a positive output is treated as 1.
   \item[Population parameter similarity] A measurement of population convergence calculated as the average pairwise cosine similarity of the population. This metric was introduced in \cite{mouret2012encouraging}.
   \item[Parent child behavior difference] A measurement of recombination disruptiveness. The behavior difference between two neural networks is calculated as the mean absolute value of the difference of output values for all inputs. The parent-child behavior difference is the mean of the behavior difference between the child and the first parent, and between the child and the second parent.
    
\end{description}

\section{Results}
\begin{figure}[ht]
  \centering
  \begin{tabular}{cc}
  \subfloat[Fitness of 8-bit parity problem]{\includegraphics[width=0.5\linewidth]{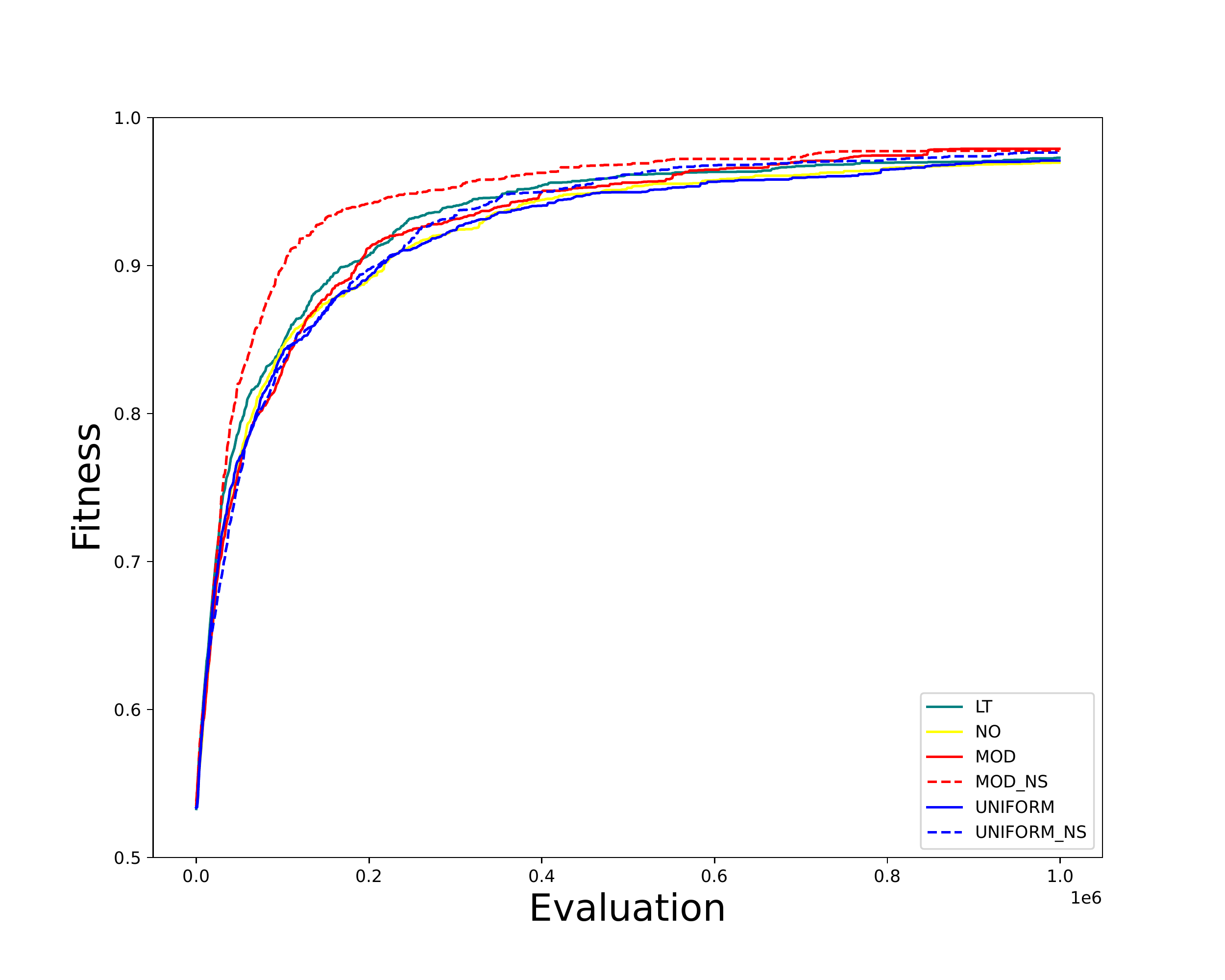}} &
  \subfloat[Fitness of 10-bit parity problem]{\includegraphics[width=0.5\linewidth]{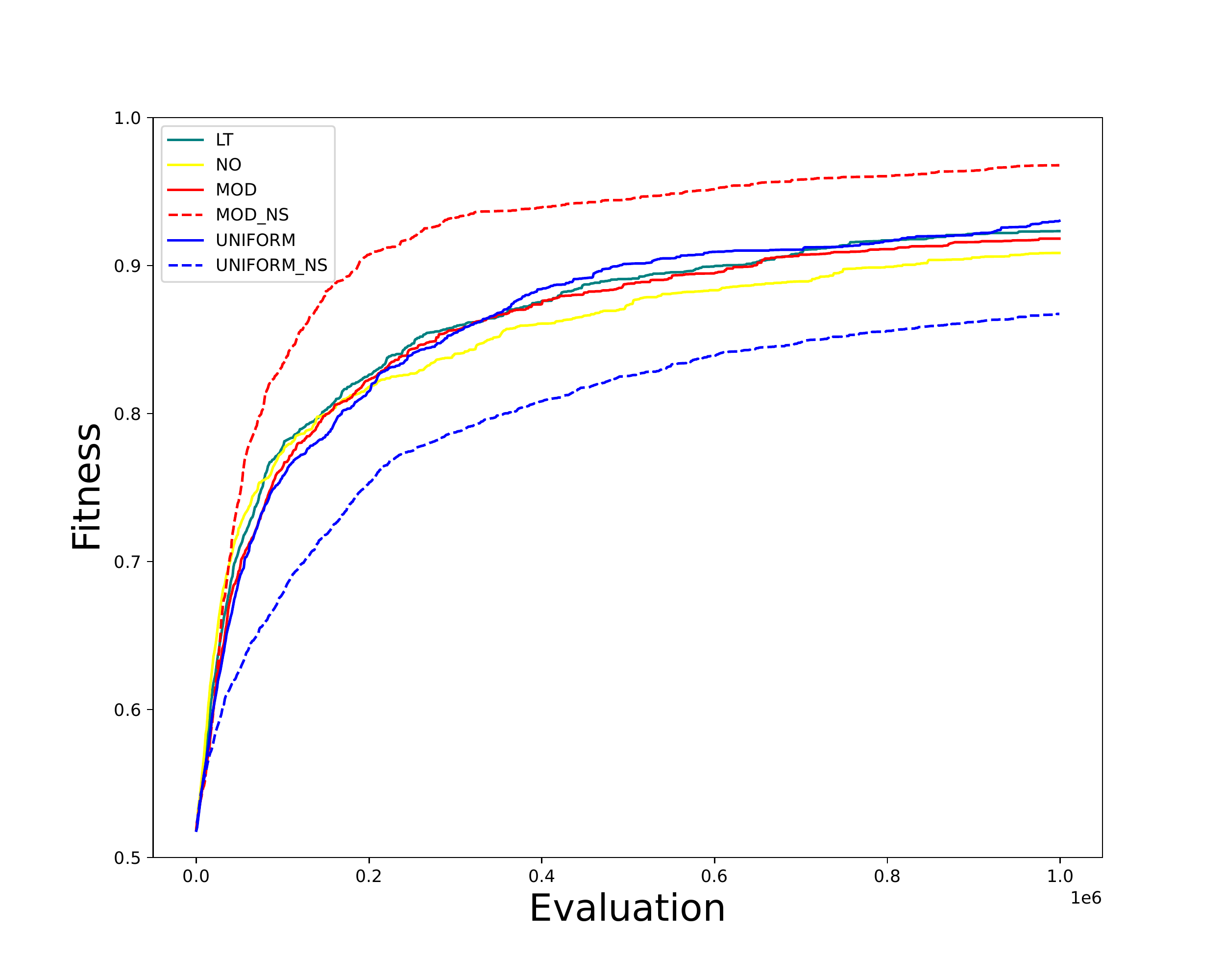}}\\
  \subfloat[\centering Average parameter cosine similarity between each pair of individual in the population of 8-bit parity problem]{\includegraphics[width=0.5\linewidth]{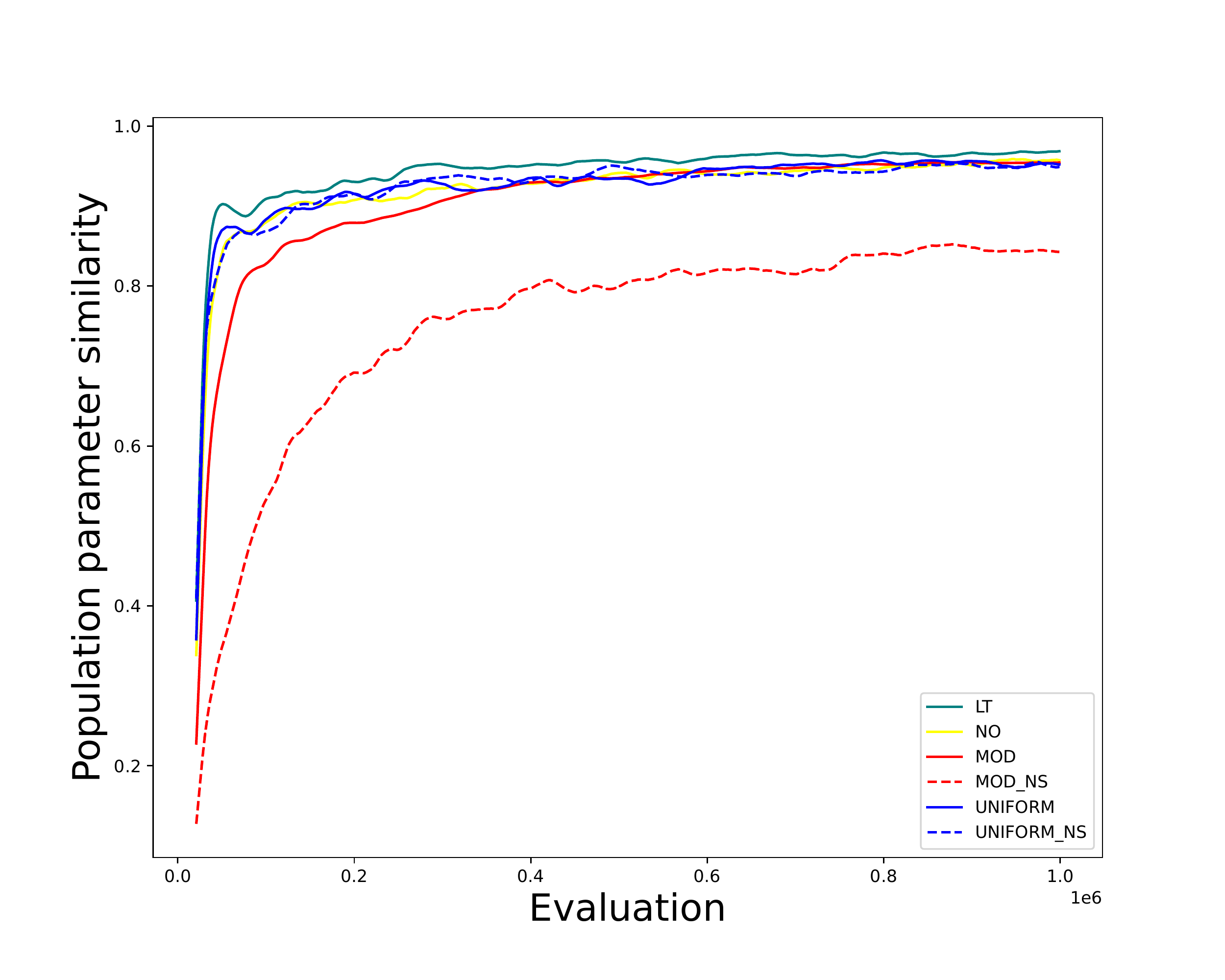}} &
  \subfloat[\centering Average parameter cosine similarity between each pair of individual in the population of 10-bit parity problem]{\includegraphics[width=0.5\linewidth]{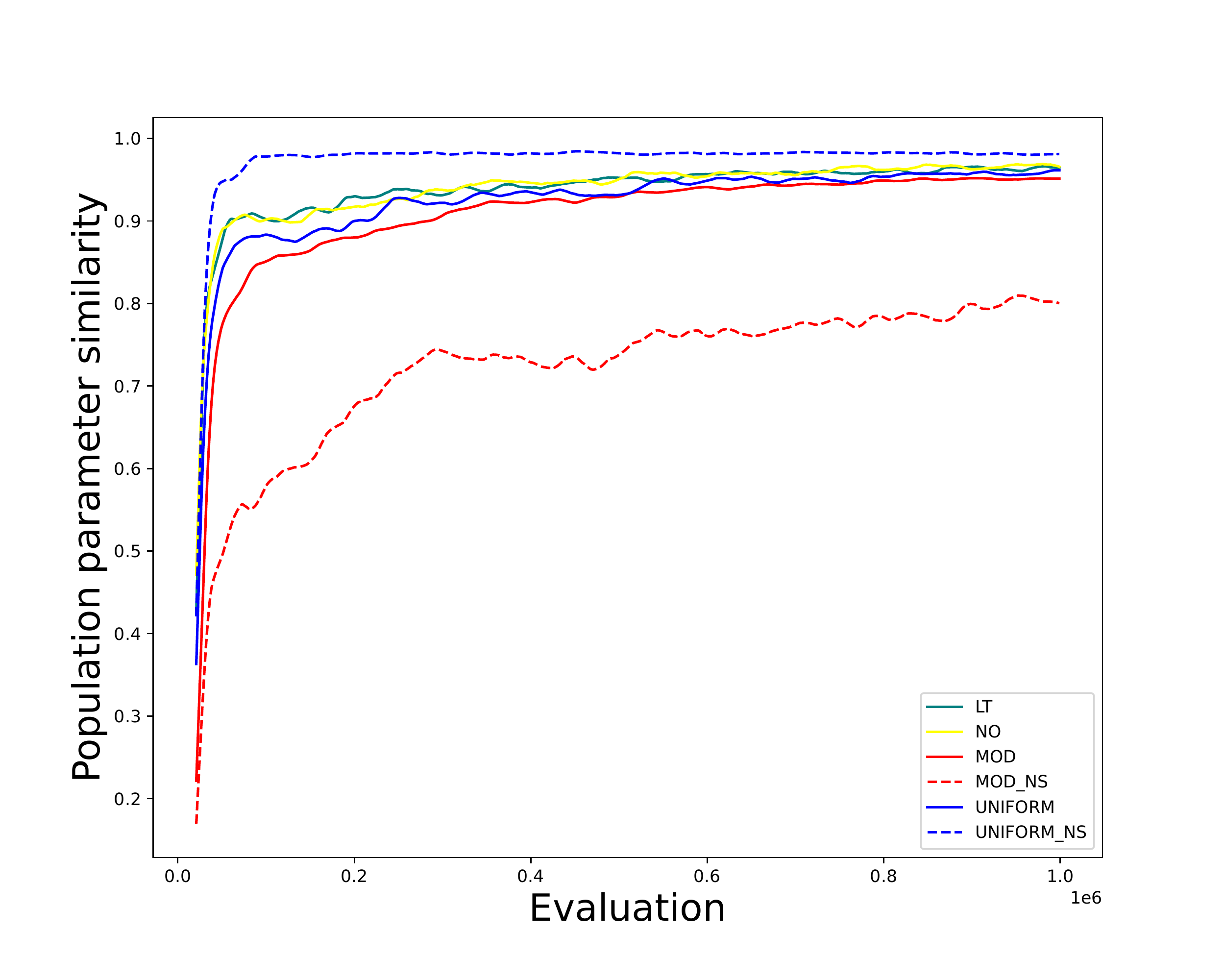}}\\
  \subfloat[\centering Average behavioral difference between parents and child of 8-bit parity problem]{\includegraphics[width=0.5\linewidth]{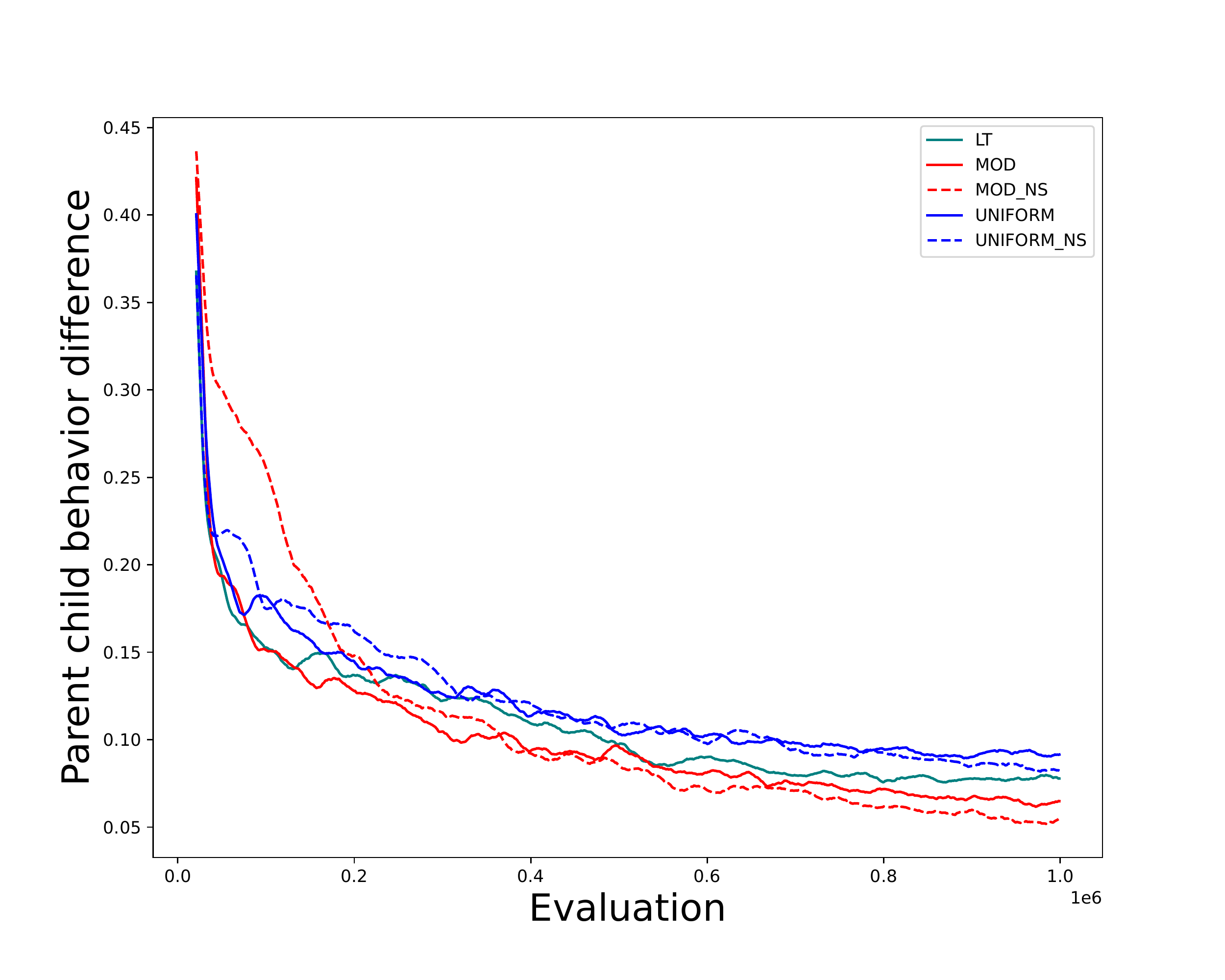}} &
  \subfloat[\centering Average behavioral difference between parents and child of 10-bit parity problem]{\includegraphics[width=0.5\linewidth]{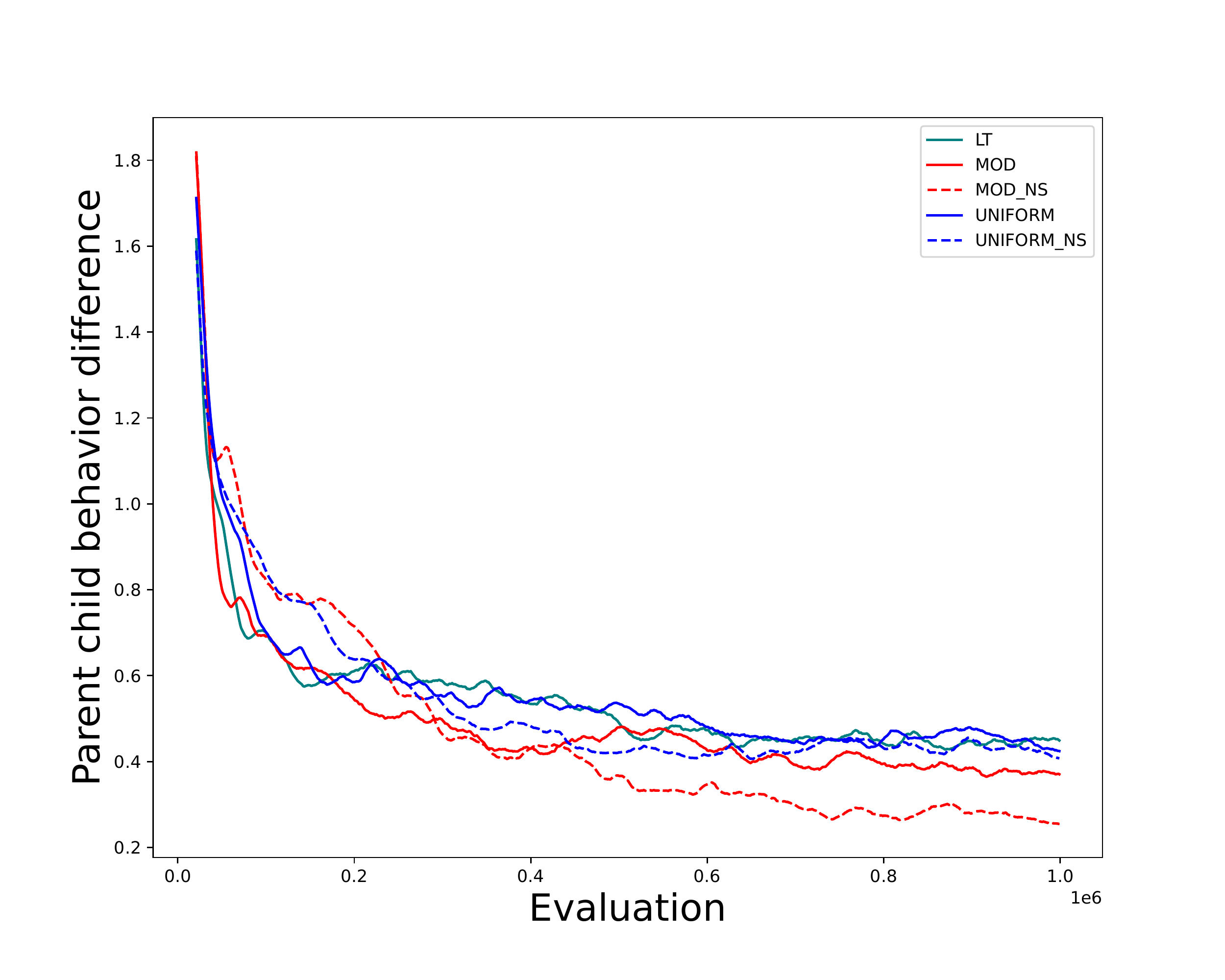}}\\
  \end{tabular}
  \caption{The results of the parity problems}
  \label{fig:result}
\end{figure}

As shown in Figure \ref{fig:result}(a)(b), for 8-bit parity MOD\_NS converges towards good solutions more quickly, though it is slightly worse than the average fitness of MOD. For 10-bit parity, MOD\_NS outperforms all other setups by a large margin. An interesting observation is that the neuron similarity operator positively impacts the performance of the modularity-based linkage model but not so much when uniform crossover is performed. In fact, for the 10-bit parity problem, NS has the exact opposite effect. To understand this, we show how the population converges in terms of parameter cosine similarity in Figure\ref{fig:result}(c)(d). In both problems, MOD\_NS has the least converged population. Our interpretation is that since neurons are rearranged according to their behavior before crossover, successful crossovers do not require the same permutation of neurons thus the diversity of neuron permutations is preserved. The NS operator however is not perfect; it effectively induces noise as the difficulty of getting the correct neuron alignment rises. There could be no 1-to-1 alignments in the first place or the complexity of the functionality overwhelms this method of simply comparing neuron activation values. In \ref{fig:result}(d) we observed extremely converged populations of UNIFORM\_NS, but not so much in \ref{fig:result}(c). A possibility is that given a disruptive recombination operator, the population of OMEAs tends to be extremely converged in order to perform non-disruptive crossover. The reason is that during Optimal Mixing, an offspring would only be accepted if outperforms its parents and this is more likely to happen when the crossover is non-disruptive, which requires both parents to be very similar when the crossover operator itself is disruptive. Population diversity can be quickly lost when all offspring are produced by extremely similar parents. This could mean the complexity of 10-bit parity is so much for the NS operator, that it starts to induce such a big noise for the already disruptive uniform crossover. In order to overcome this, the population has to be extremely converged to negate the noise so non-disruptive crossovers can be performed so that the offspring can be accepted. However the NS operator is helpful on both problems for the modularity-based linkage model. We hypothesize that this is because NS works by reducing effective search space by achieving permutation invariance in terms of neuron orders and if the population is already very converged, the permutation problem would not exist anyway thus the contribution of NS would be purely negative. That is, solving the permutation problem can barely help with recombination if high disruption exists to push toward an early-converged population. According to \ref{fig:result}(c)(d), MOD has significantly less-converged populations thus it benefits from NS more, despite it causing some degree of disruption to the recombination.

In Figure \ref{fig:result}(e)(f), we measure how differently an offspring would behave from its parents to understand how disruptive each recombination operator can be. While MOD\_NS has the highest difference in some of the early stages, it gradually achieves the lowest average behavioral difference by the end for both problems while maintaining high diversity in the parameter space. Modularity-based linkage model aims to preserve substructures during recombination, but in the early stage of evolution, substructures might not yet emerge, or, the overall structures are so diverse in the population, attempting to swap substructures between individuals is deemed to be disruptive. This combined with the fact that NS is also disruptive while there is not good alignment between neurons from both parents, explains why MOD\_NS performs disruptive recombination in the early stage. As substructures emerge in the process, MOD\_NS and MOD become less and less disruptive compared with other settings. 

\begin{figure*}[ht]
  \centering
  \begin{tabular}{cccc}
  \subfloat[]{\includegraphics[width=0.25\linewidth]{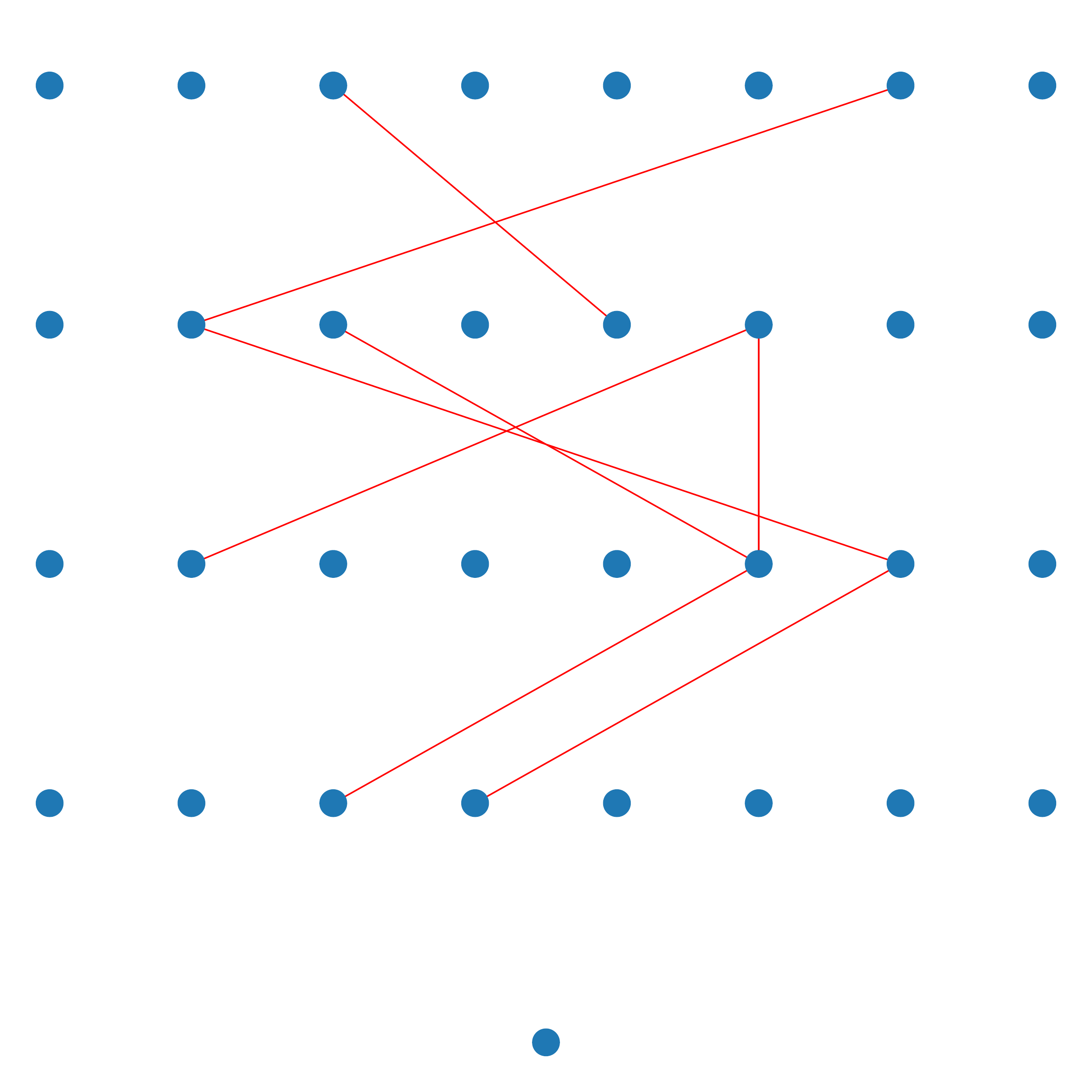}} &
  \subfloat[]{\includegraphics[width=0.25\linewidth]{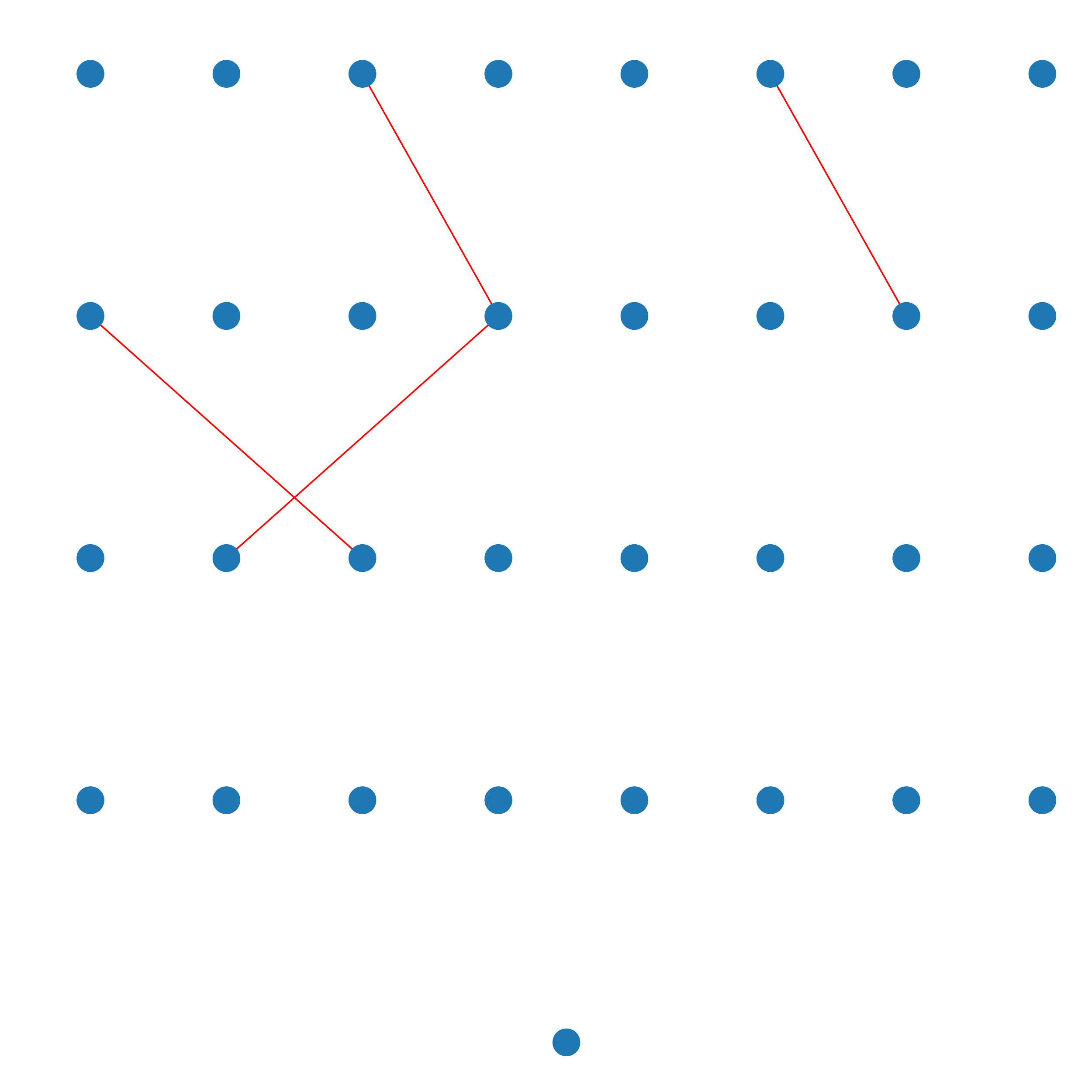}} &
  \subfloat[]{\includegraphics[width=0.25\linewidth]{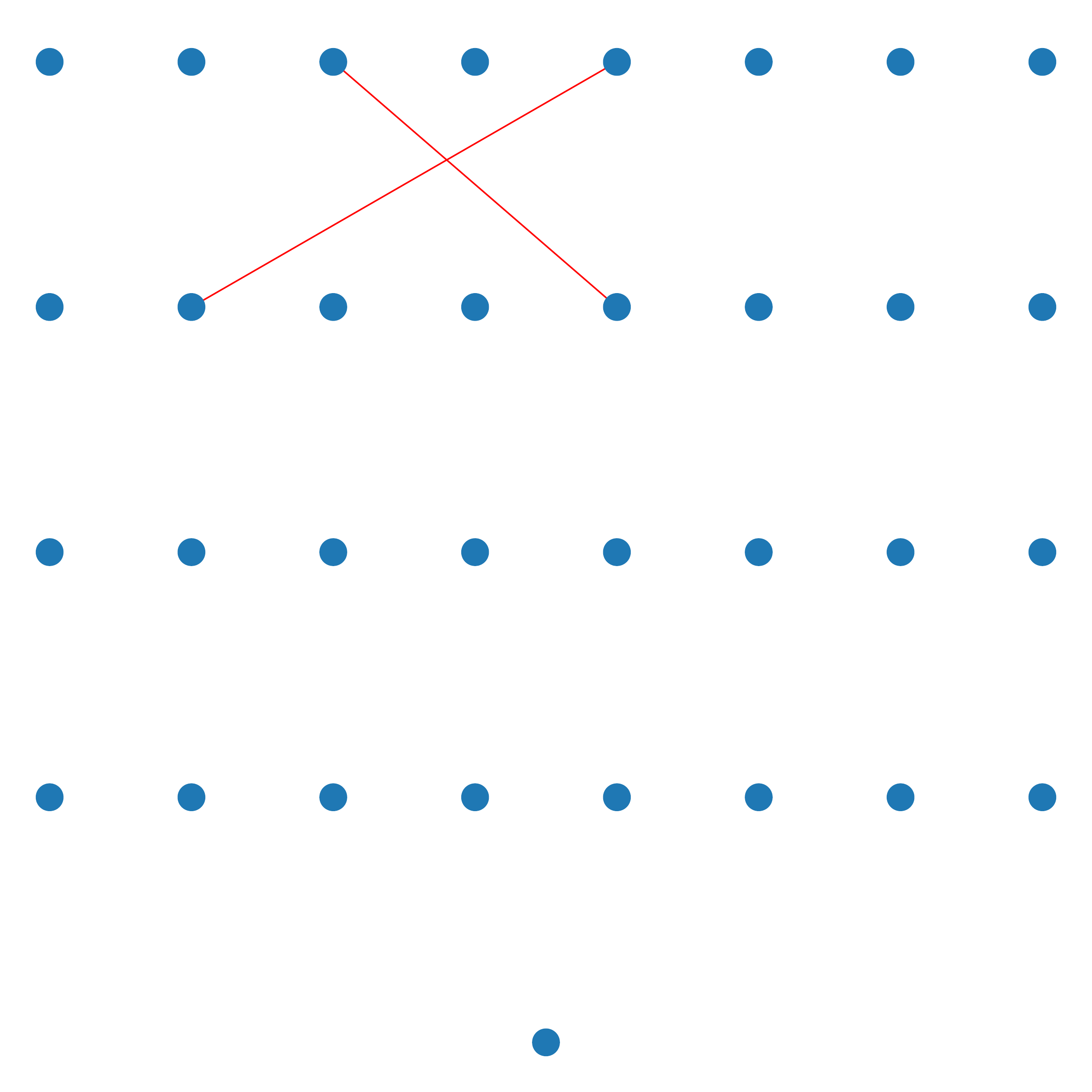}} &
  \subfloat[]{\includegraphics[width=0.25\linewidth]{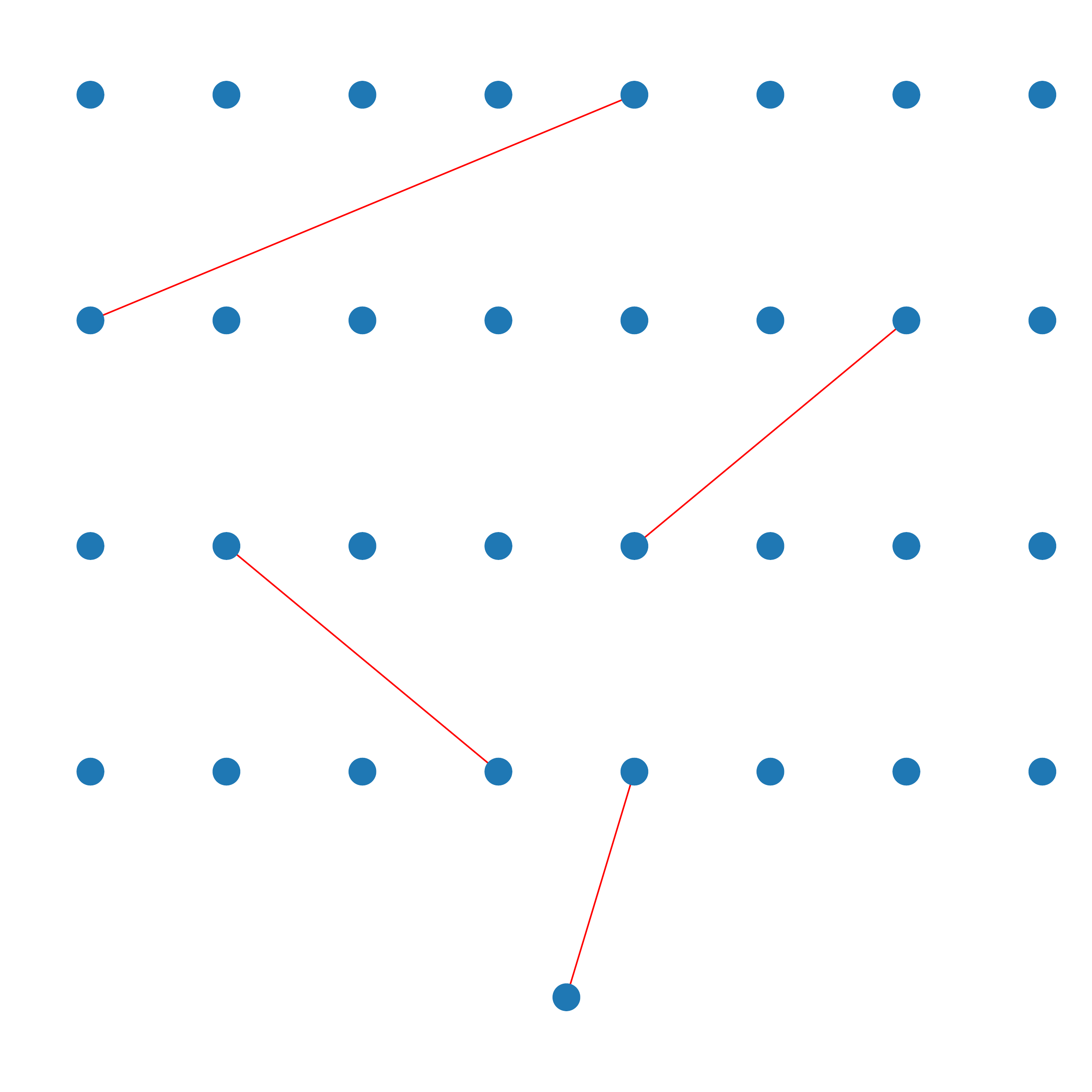}} \\
  \subfloat[]{\includegraphics[width=0.25\linewidth]{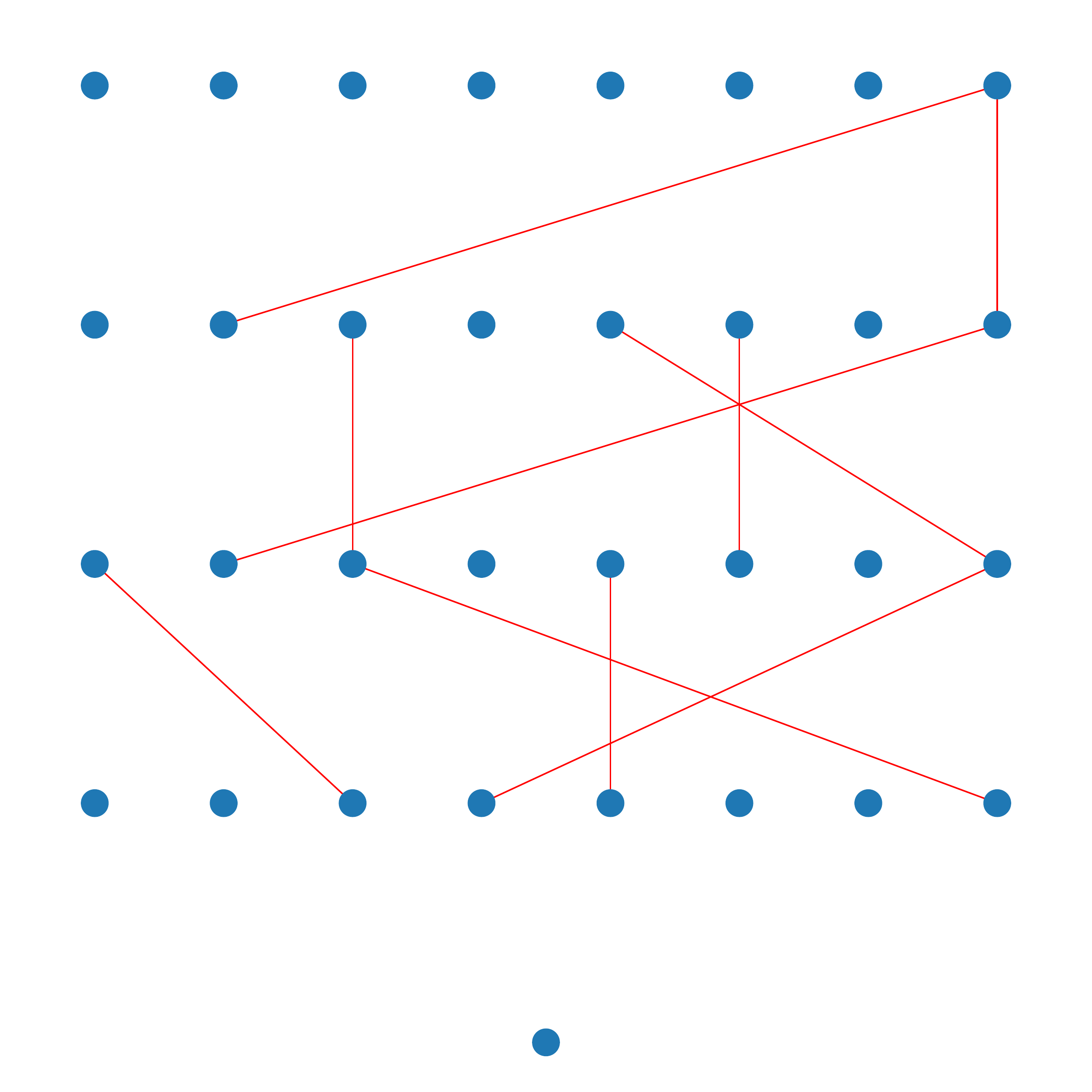}} &
  \subfloat[]{\includegraphics[width=0.25\linewidth]{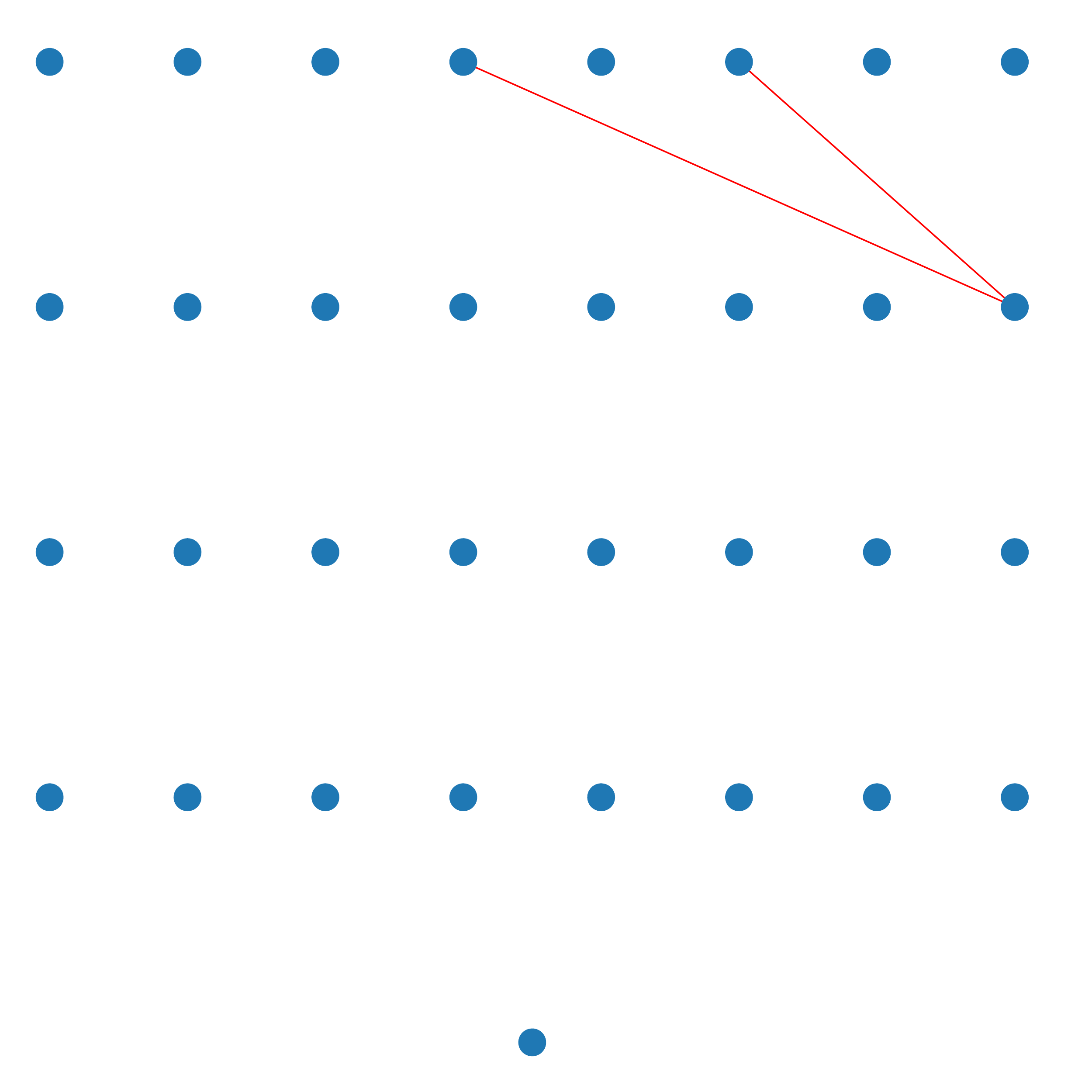}} &
  \subfloat[]{\includegraphics[width=0.25\linewidth]{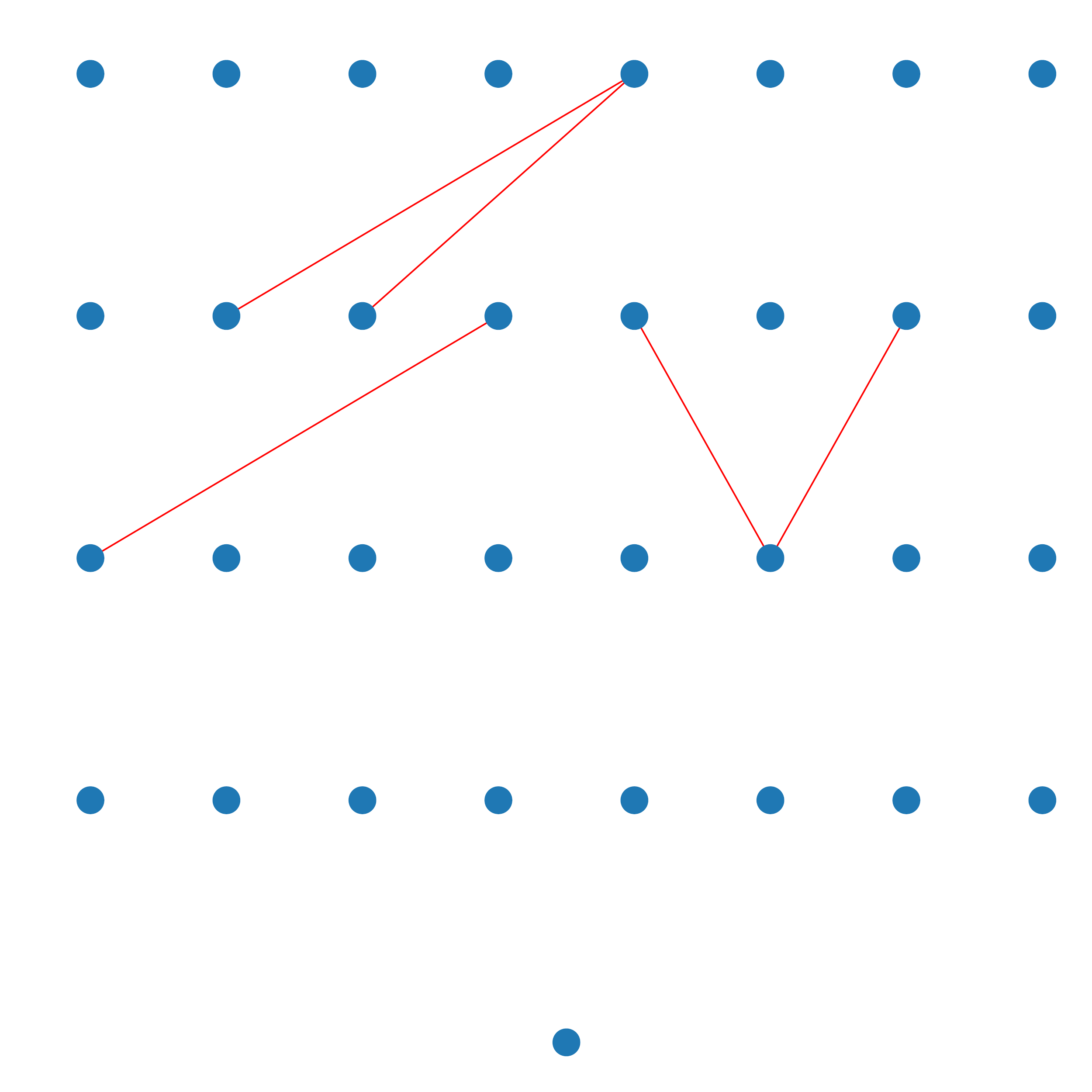}} &
  \subfloat[]{\includegraphics[width=0.25\linewidth]{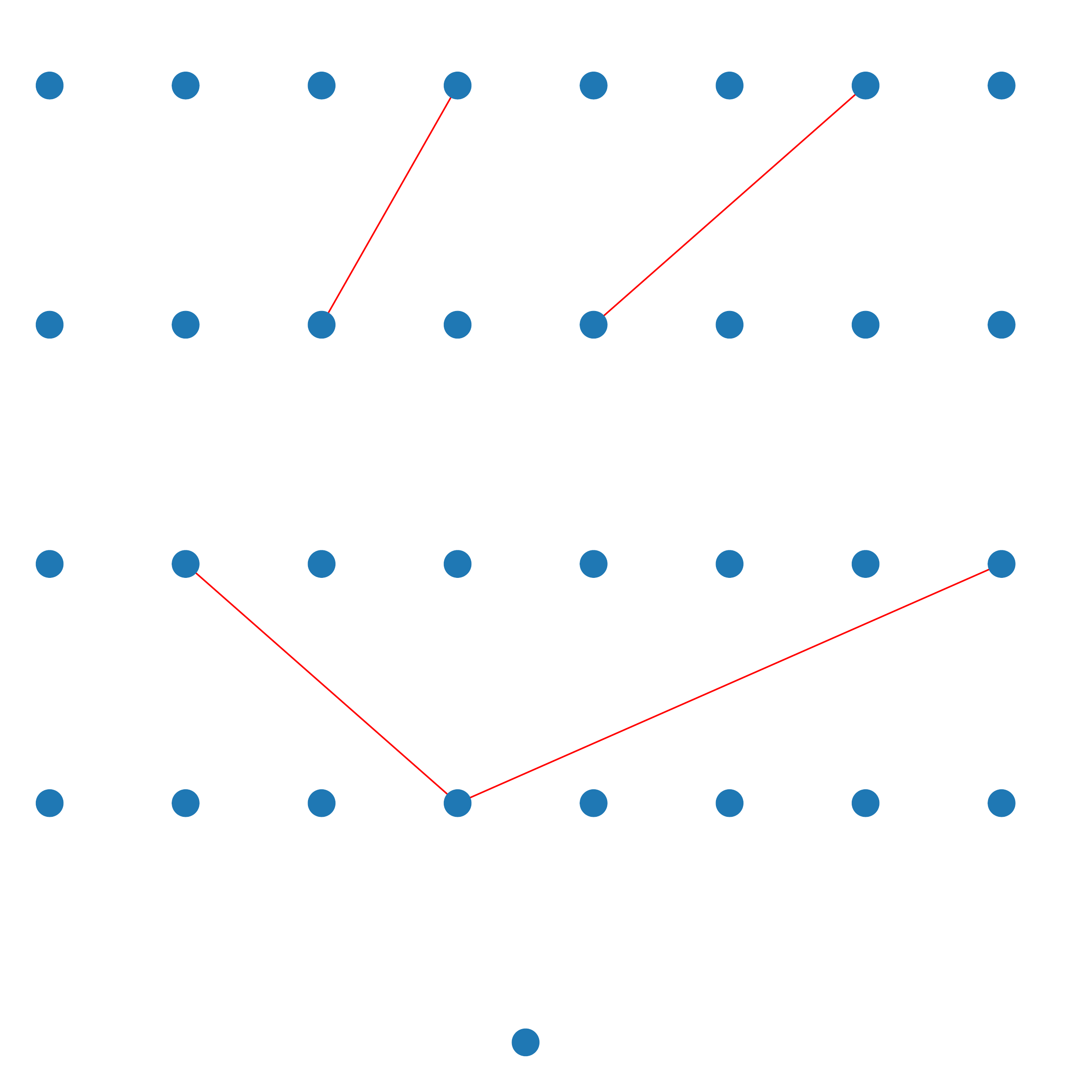}} \\
  \subfloat[]{\includegraphics[width=0.25\linewidth]{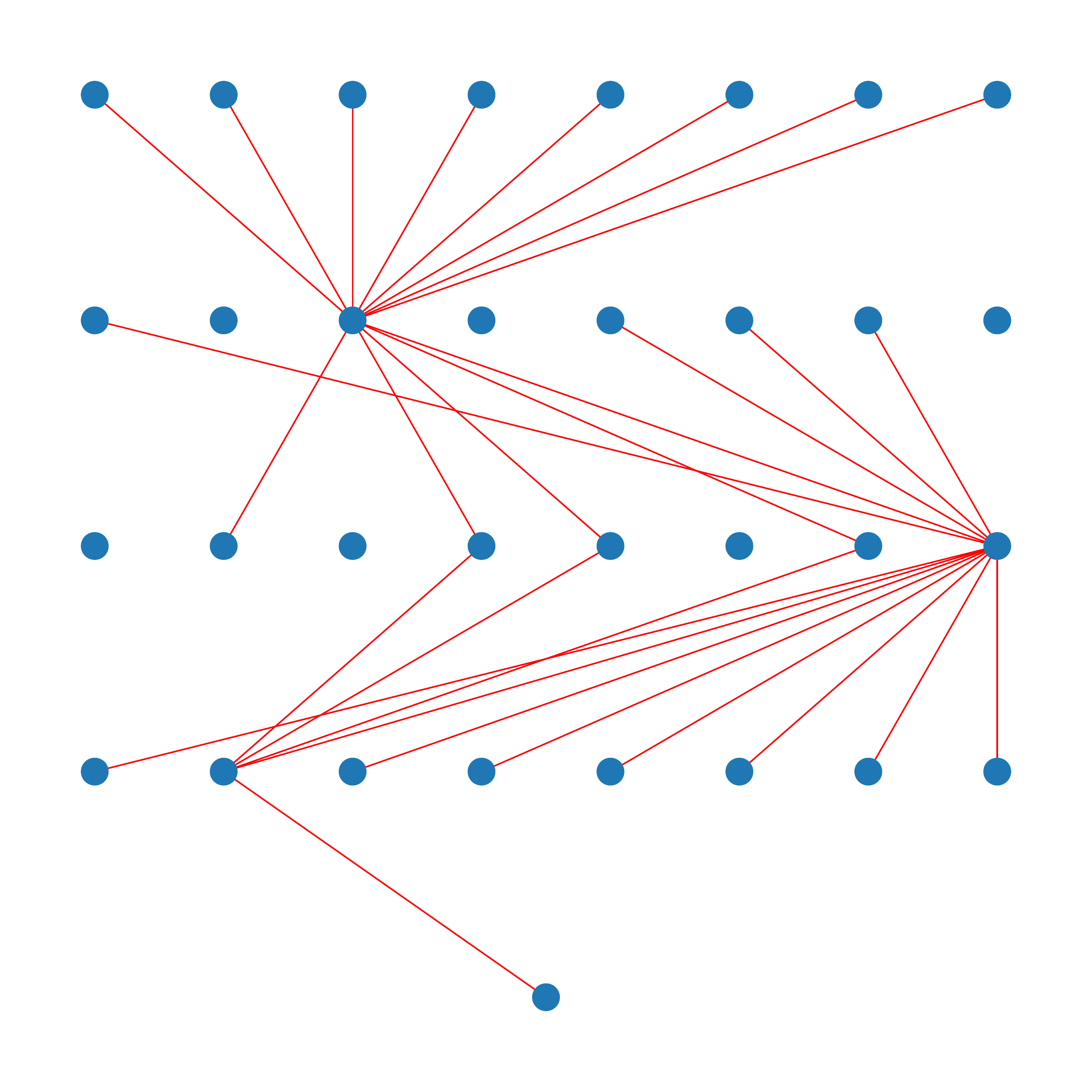}} &
  \subfloat[]{\includegraphics[width=0.25\linewidth]{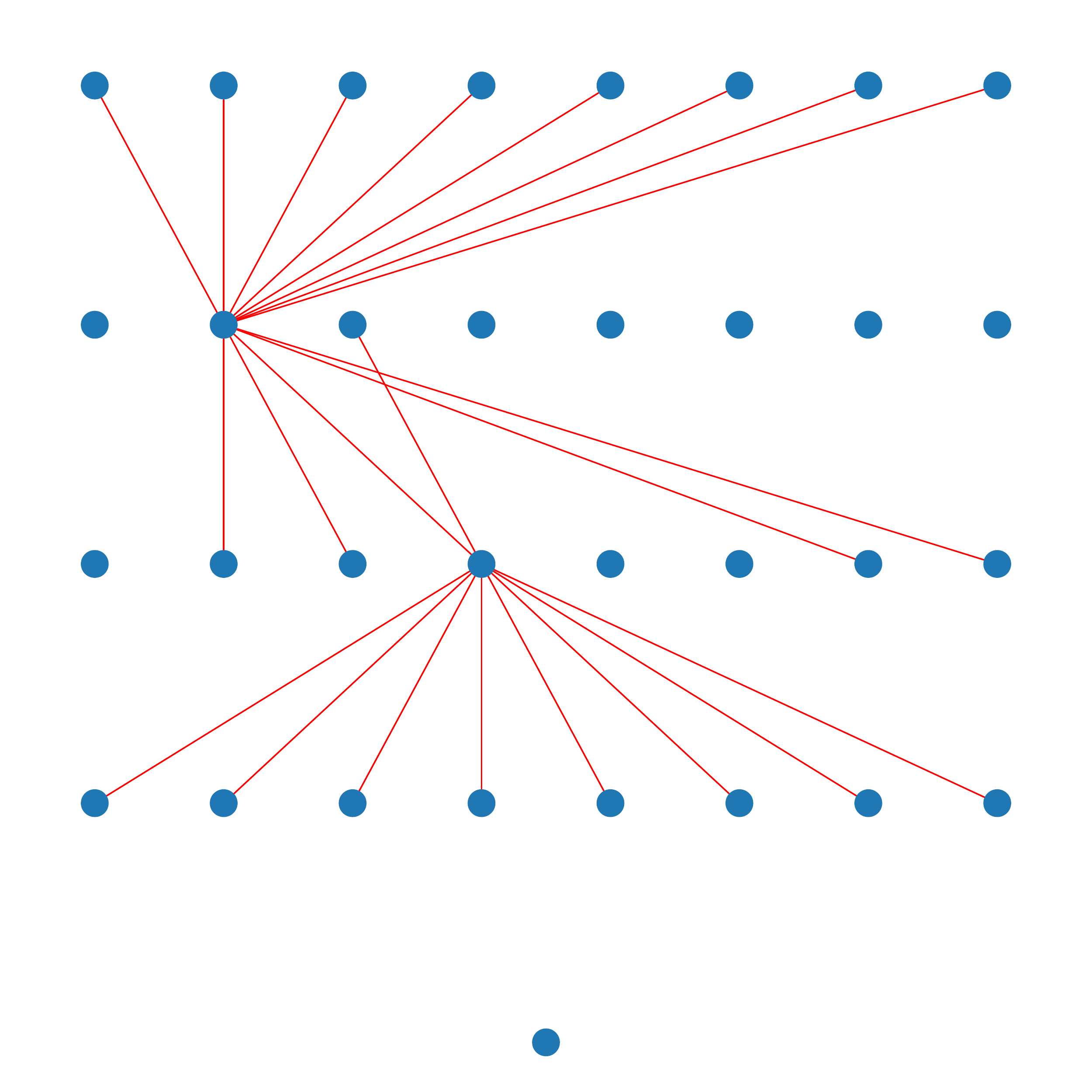}} &
  \subfloat[]{\includegraphics[width=0.25\linewidth]{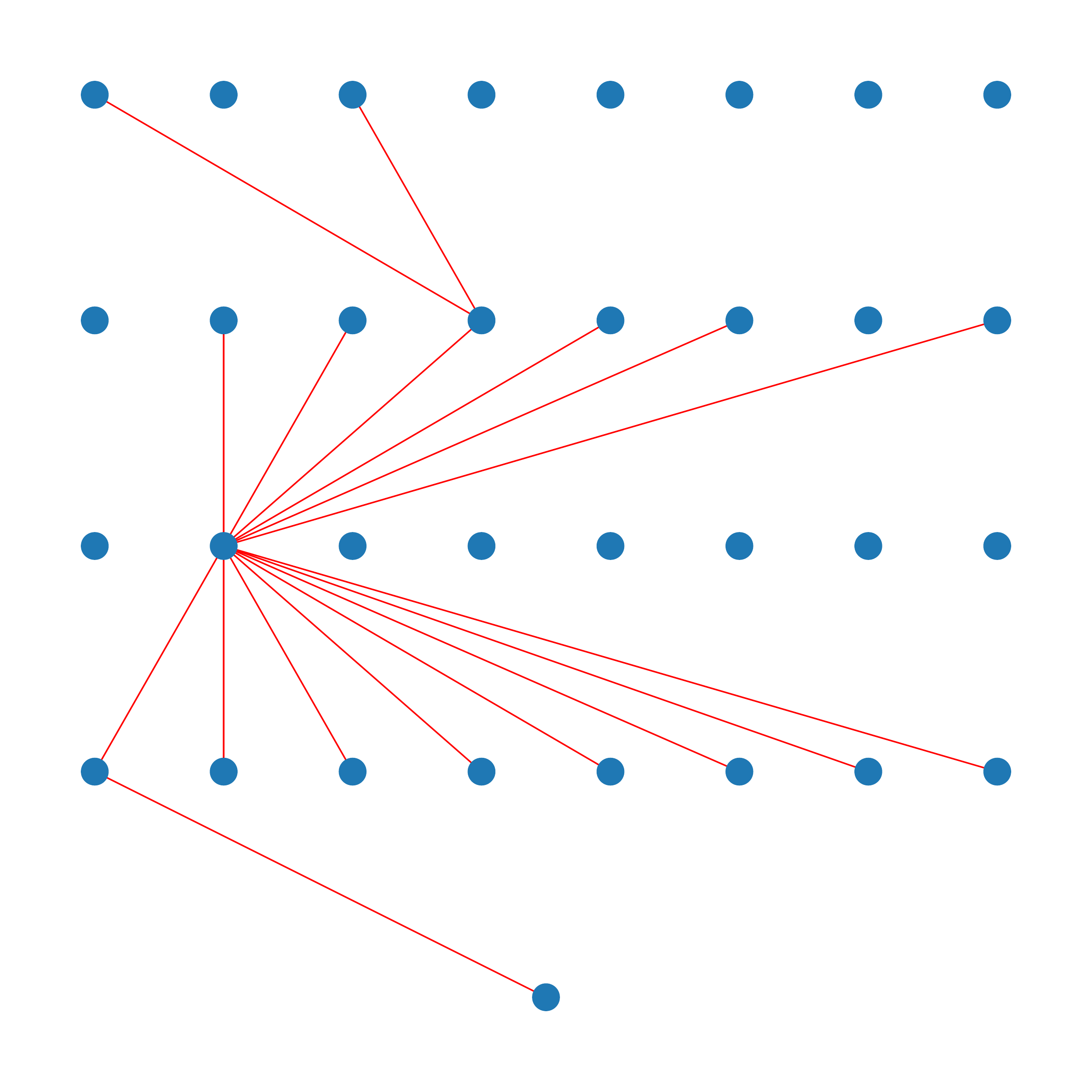}} &
  \subfloat[]{\includegraphics[width=0.25\linewidth]{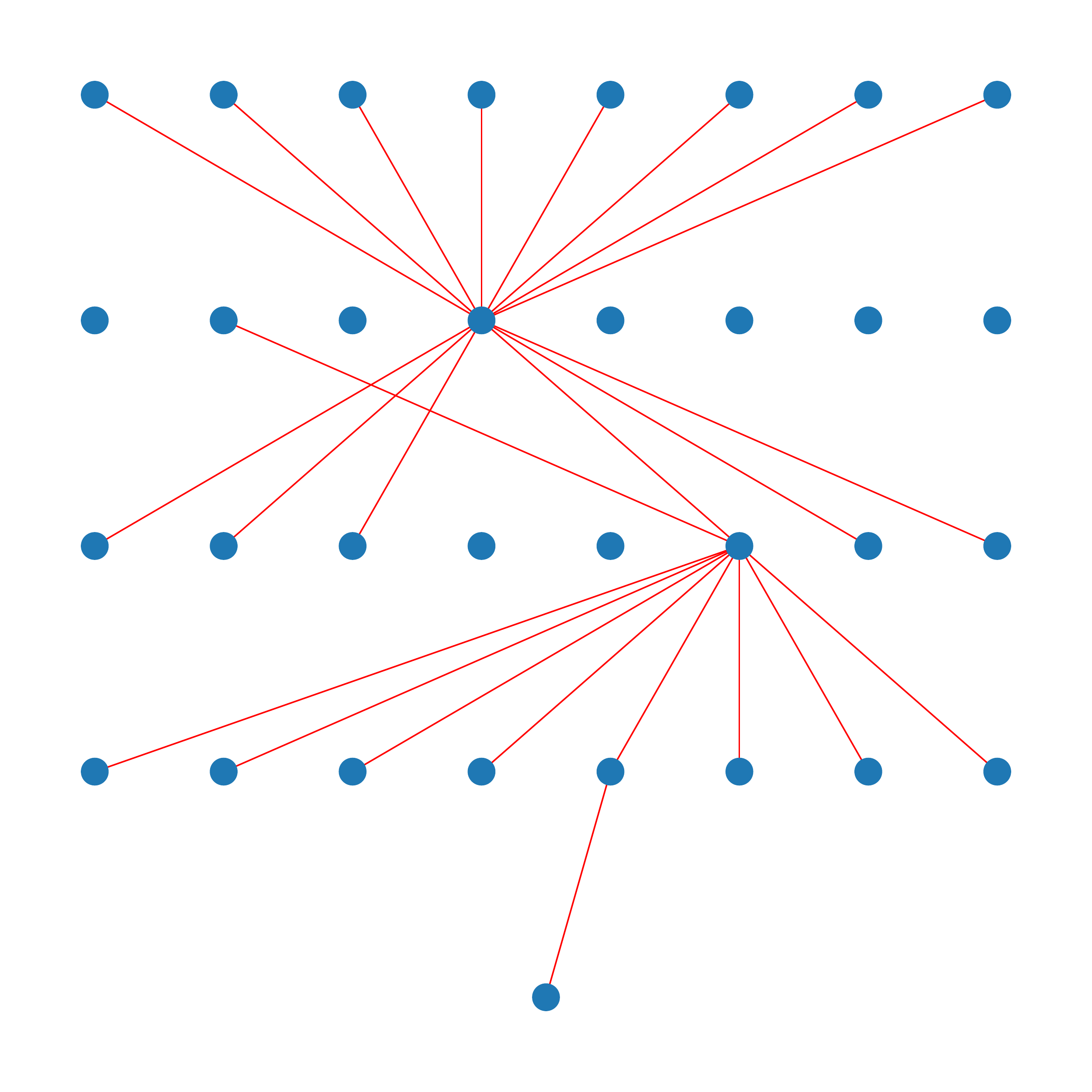}} \\
  \subfloat[]{\includegraphics[width=0.25\linewidth]{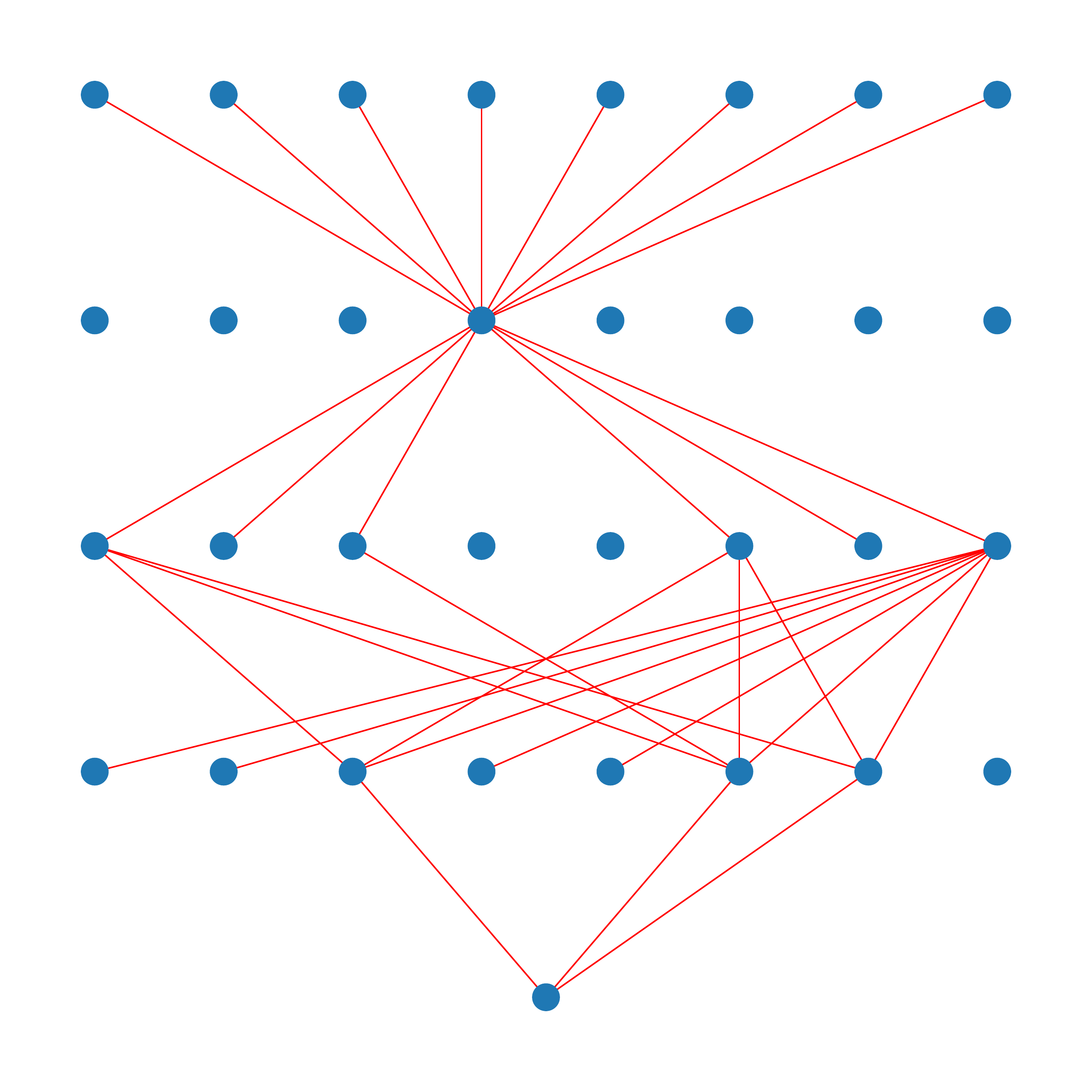}} &
  \subfloat[]{\includegraphics[width=0.25\linewidth]{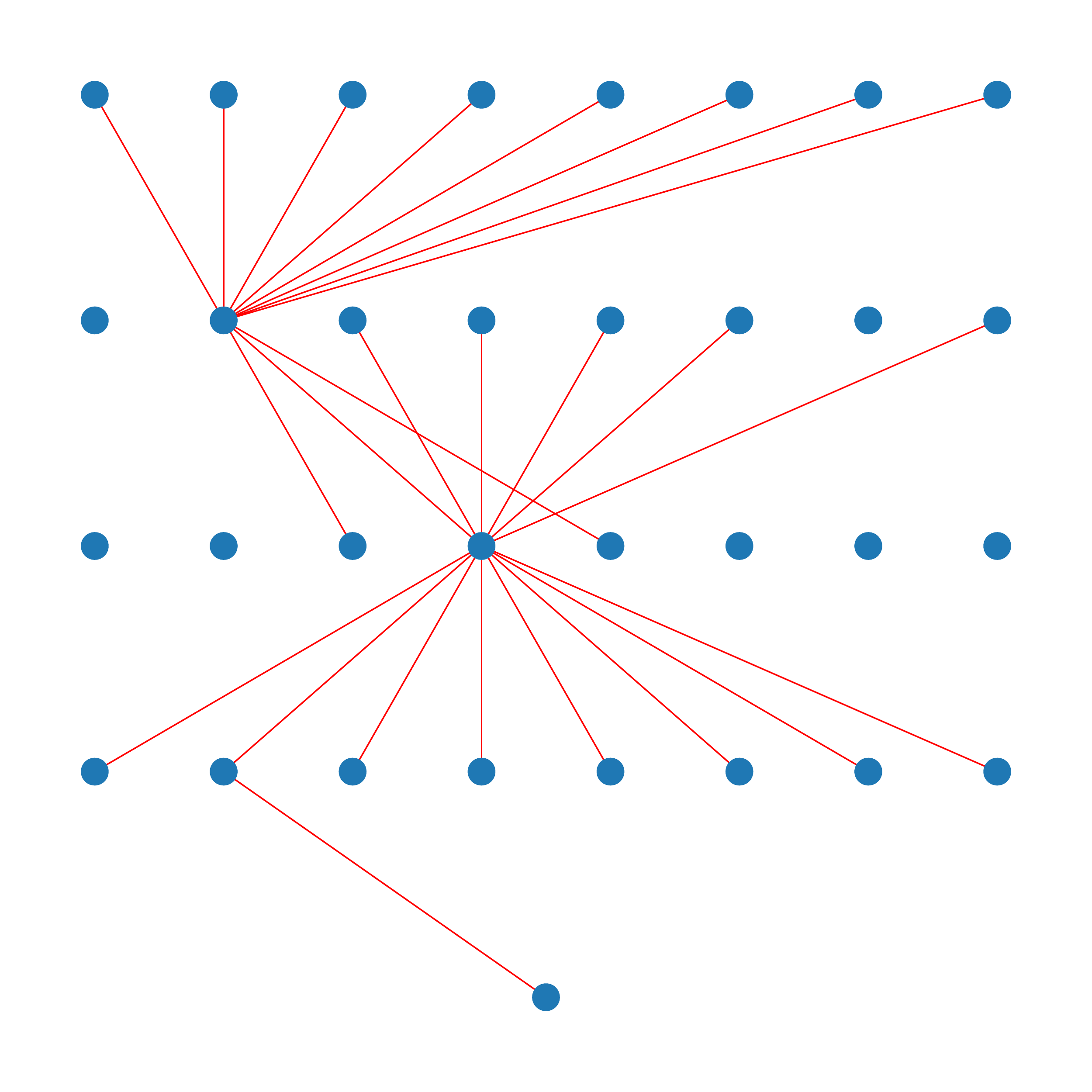}} &
  \subfloat[]{\includegraphics[width=0.25\linewidth]{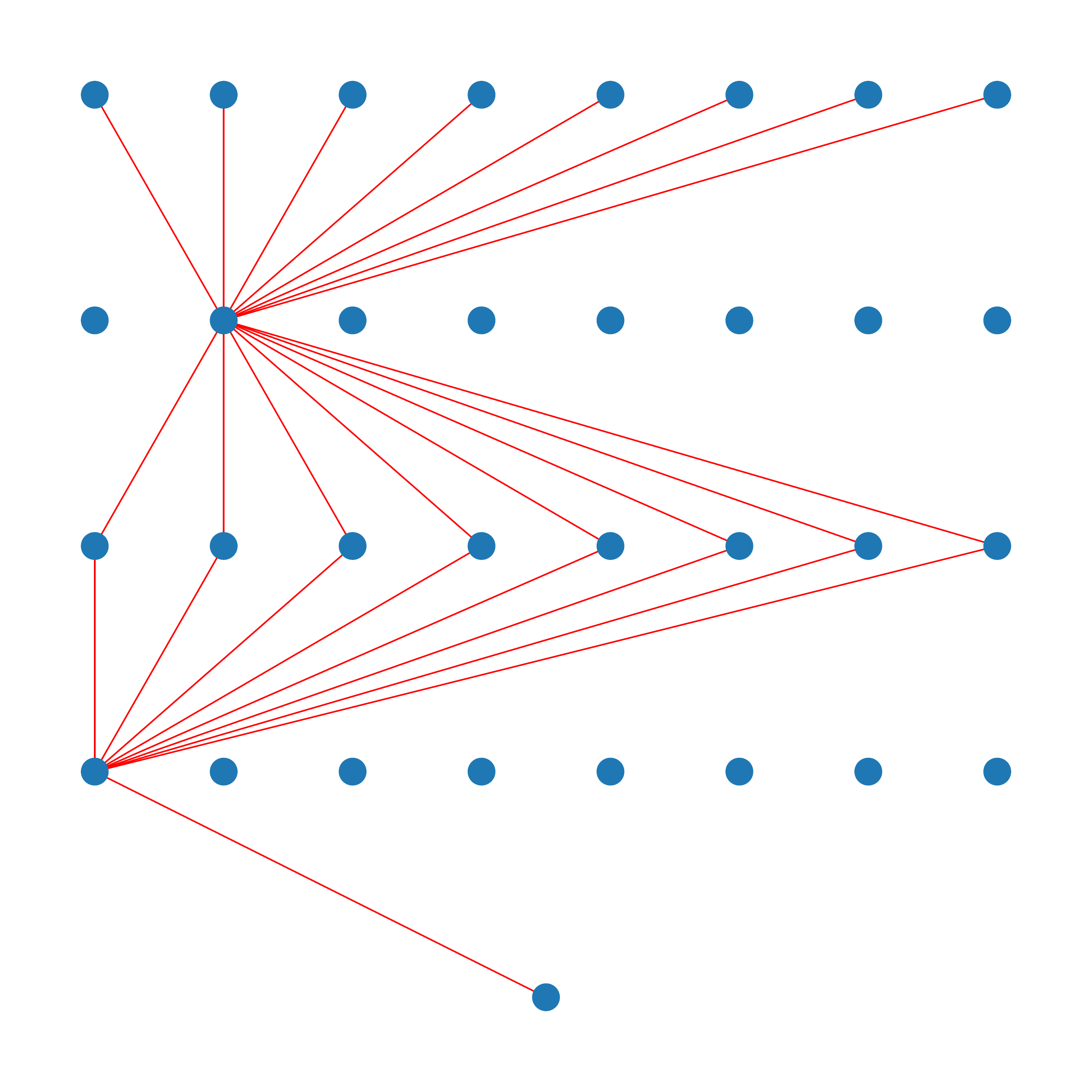}} &
  \subfloat[]{\includegraphics[width=0.25\linewidth]{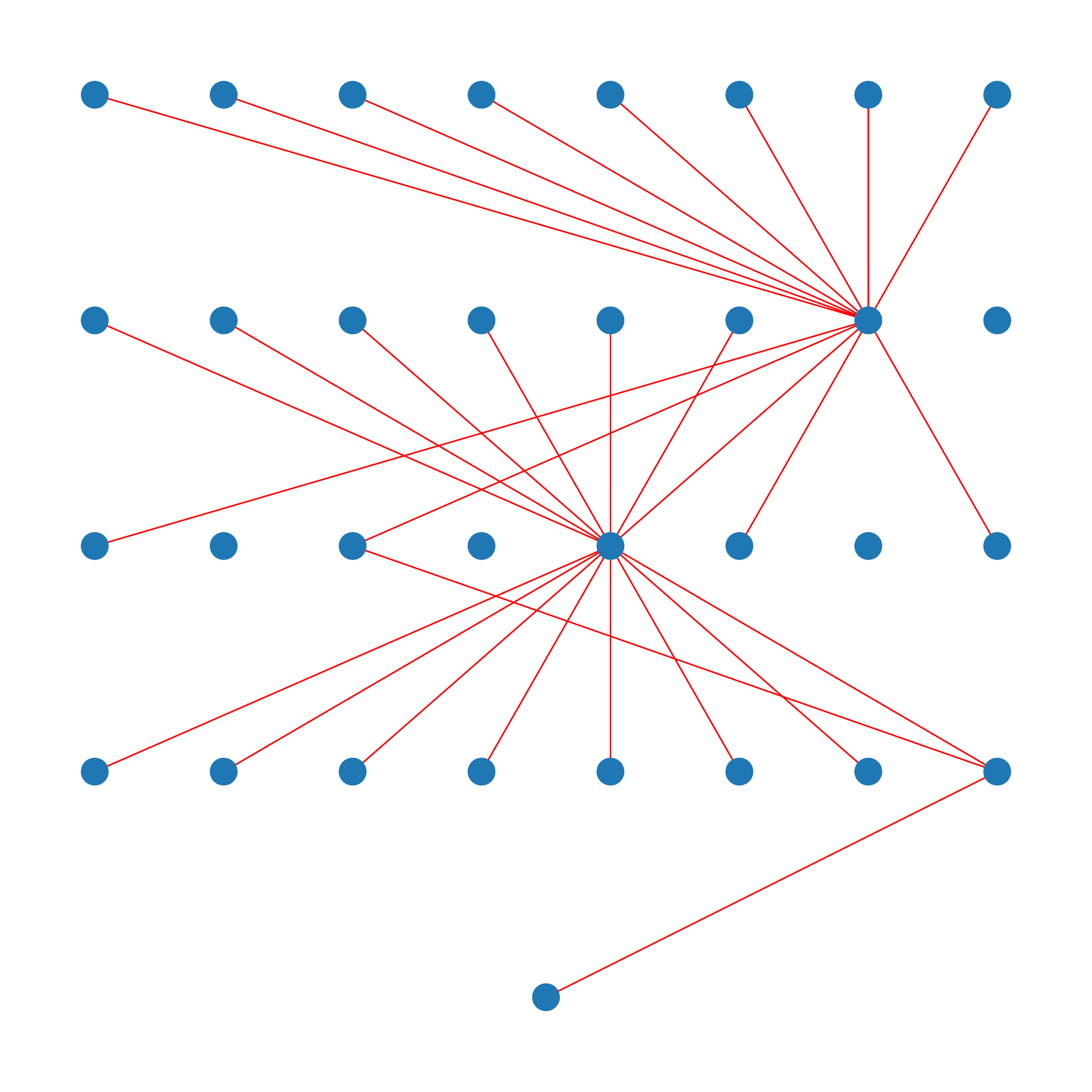}} \\
  \end{tabular}
  \caption{Linkages that produced successful recombination randomly chosen from solutions of the 8-bit parity problem. (a) to (h): Linkage Tree; (i) to (p): Modularity-based}
  \label{fig:linkage}
\end{figure*}

In OM with FOS an offspring is only accepted when crossover with the traversed FOS element results in a performance increase. A big factor of recombination disruptiveness is the amount of genetic material exchanged. Swapping only a tiny part of the neural network should not have as much of an impact on its functionality compared to swapping half of the network. This is also a reason for uniform crossover being such a disruptive operator since its crossover rate is fixed at 0.5. A reasonable deduction is that the size of accepted subsets tells us the quality of the linkage being modeled by the FOS, as a well-separated subset should be functionally independent regardless of its size. Table.\ref{Tab:crossrate} shows that MOD and MOD\_NS have a much higher average rate of genetic material exchange compared to LT, meaning much larger subsets being accepted during OM. Note that when the exchange rate is greater than 0.5, (1 - the exchange rate) is considered to be the crossover rate instead. To further examine the quality of the FOS, Figure.\ref{fig:linkage} shows some of the accepted subsets randomly selected from the experiments. Note that when $|\mathcal F_{i}|>\frac{|S|}{2}$ we show $S-|\mathcal F_{i}|$ instead since it is hard to see anything when the FOS element contains almost all of the parameters. Although being small in size and containing disjoint connections, LT identifies connected parameters with a higher probability than random, even without prior knowledge of neural networks. A big problem, however, is that LT learns either very large linkage or very small linkage as the population converges. This is expected since the optimization of neural network parameters is such an inseparable problem, that every weight is dependent on each other to some degree. This intractable noise would make it extremely hard to find relatively independent functional substructures without any domain knowledge. The low crossover rate also explains why the convergence of the parameters of the population of LT is even more severe compared to UNIFORM where a disruptive crossover is performed. With our model specifically designed for fully connected feed-forward neural networks, by focusing on first-order interactions only, this noise can be overcome at the price of ignoring more complex interactions. This is reflected by the fact that linkages found are always well-connected as shown in Figure.\ref{fig:linkage}, which means the inter-dependency within the subsets is being respected. Combined with the fact that MOD and MOD\_NS cause the least amount of behavioral difference during recombination, we conclude that our method finds linkages with good quality.

%\begin{center}
%\begin{table}
%\begin{tabular}{ |c|c|c| } 
% \hline
%  & 8-bit & 10-bit \\ 
% \hline
% LT & 10.03 & 10.58 \\ 
% NO & 9.59 & 9.77 \\ 
% MOD & 10.61 & 10.26 \\ 
% MOD\_NS & 10.19 & 10.16 \\
% UNIFORM & 9.64 & 10.12 \\ 
% UNIFORM\_NS & 9.63 & 9.89 \\
% \hline
%\end{tabular}
%\caption{Average successful reproduction rate in percentage}
%\end{table}
%\end{center}

\begin{center}
\begin{table}
\begin{tabular}{ |c|c|c| } 
 \hline
  & 8-bit & 10-bit \\ 
 \hline
 LT & 1.14 & 0.57 \\ 
 MOD & 9.90 & 5.18 \\ 
 MOD\_NS & 8.21 & 5.23 \\
 \hline
\end{tabular}
\caption{Average cross rate of successful crossover in percentage}
\label{Tab:crossrate}
\end{table}
\end{center}

\section{Conclusion}
This paper has identified two main sources of disruption of recombination between fully connected feed-forward neural networks with direct encoding, the permutation problem and the substructure disruption problem. A modularity-based linkage model was proposed to solve the latter problem while the neuron similarity operator, inspired by\cite{dragoni2014simba}, was applied in combination, to reduce the former problem. We have shown that this combination outperforms the Linkage Tree model, as it reaches higher fitness in fewer function evaluations on the challenging parity problem. It produces offspring that are functionally close to their parents while maintaining a diverse population in the parameter space. it is also shown that the actual linkage information learned by our model is of much higher quality compared to the linkage learned by LT as the size of the accepted linkage is much larger. Finally, we showed that solving either of the permutation problem or the substructure disruption problem is not enough for efficient recombination, but both need to be assessed at the same time.

\section{Future Work}
While revealing the important factors toward efficient recombination between fully connected neural networks, the problems are only addressed by crude estimations, despite producing good results. It may well be possible to make further improvements if more sophisticated and accurate methods are applied, but computational demands could become a problem. Future work on better dependency estimation methods and better solutions to the permutation problem is an open area of research towards the feasibility of scaling neuroevolution to large, modular solutions for challenging problems.

%\begin{acks}

%\end{acks}

\bibliographystyle{ACM-Reference-Format}
\bibliography{sample-base}

\appendix

\end{document}